\pdfoutput=1
\PassOptionsToPackage{table}{xcolor}

\documentclass[11pt]{article}
\usepackage{xcolor}

\usepackage[]{acl}

\usepackage{times}
\usepackage{latexsym}

\usepackage[T1]{fontenc}
\usepackage{csquotes}
\usepackage{array}

\usepackage{multirow}
\usepackage{booktabs,tabularx,makecell}
\newcolumntype{L}{>{\raggedright\arraybackslash}X}
\newcolumntype{C}{>{\centering\arraybackslash}X}
\definecolor{rowA}{RGB}{245,245,245} 
\definecolor{rowB}{RGB}{230,230,230} 

\PassOptionsToPackage{table}{xcolor}

\usepackage{ragged2e}
\usepackage{array}

\usepackage{siunitx}
\usepackage[most]{tcolorbox}
\tcbset{colback=gray!3, colframe=gray!50, boxrule=0.4pt, arc=1pt}

\newtcblisting{promptbox}{
listing only,
breakable,
title={},
left=4pt, right=4pt, top=4pt, bottom=4pt,
listing options={
basicstyle=\ttfamily\footnotesize,
breaklines=true, breakatwhitespace=false,
columns=fullflexible, keepspaces=true,
showstringspaces=false
}
}

\usepackage[utf8]{inputenc}

\usepackage{microtype}

\usepackage{inconsolata}
\usepackage{float}
\usepackage{graphicx}
\usepackage{enumitem}

\title{DRInQ: Evaluating Conversational Implicature \\with Controlled Context Variation}

\author{
  Hirona Jacqueline Arai\and Xiang Ren\\
  University of Southern California \\
  \texttt{\{hjarai, xiangren\}@usc.edu}
}

\usepackage{float}
\begin{document}

\maketitle
\begin{abstract}

Human conversation relies heavily on \textit{conversational implicature}, in which speakers convey meanings that are suggested rather than explicitly stated. Although recent large language models (LLMs) exhibit strong conversational fluency, they remain unreliable when interpretation depends on reasoning that integrates social and contextual cues, a process rarely articulated in text.
We introduce \textbf{DRinQ}, a benchmark for evaluating pragmatic reasoning about conversational implicature in question utterances, designed to isolate pragmatic variation while holding each question’s surface form fixed.
 
To support scalable evaluation, we propose a semi-automated pipeline that produces question-context-interpretation instances with systematic variation.
Across evaluations, we find a consistent generation-inference asymmetry: while state-of-the-art models can generate plausible pragmatic scenarios when guided, they often fail to recover the intended implication at inference time. For smaller models, structured prompting improves alignment with human judgments. A comparative writing study further reveals complementary strengths: human authors tend to produce safer, predictable contexts, whereas models generate varied scenarios with interpretations that sometimes exceed contextual support. 
These findings highlight persistent challenges in modeling conversational implicature and motivate more context-sensitive evaluation frameworks.

\end{abstract}

\section{Introduction}
\begin{figure}[h]
    \centering
    \includegraphics[width = 0.9\linewidth]{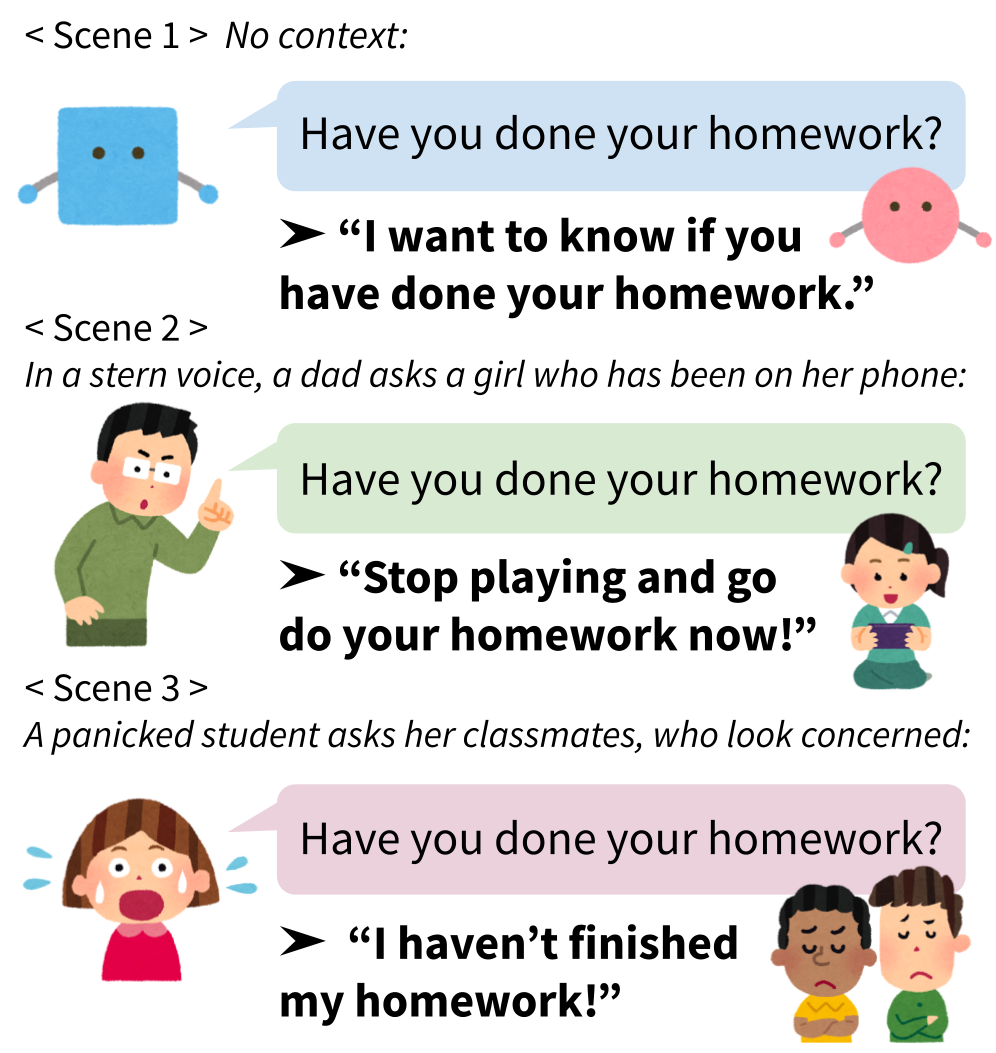}
    \caption{How conversational context reshapes implicature. The main takeaway from the utterance shifts from a genuine inquiry (scene 1) to a rhetorical rebuke (scene 2) or a desperate plea (scene 3).}
    \label{fig:intro}
\end{figure}

\footnotetext{Dataset available at \href{https://github.com/hjarai/drinq}{https://github.com/hjarai/drinq}}

\begin{figure*}[h]
    \centering
    \includegraphics[width=\textwidth]{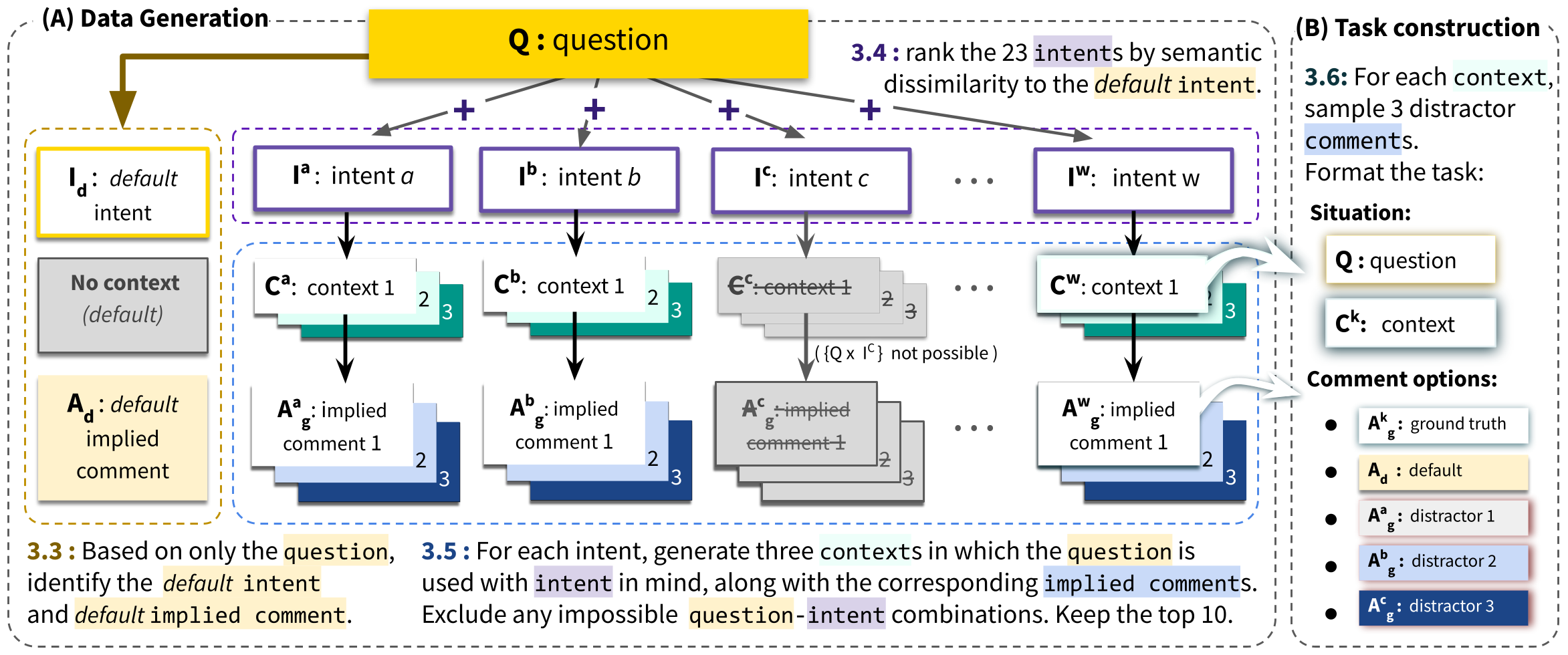}
    \caption{Pipeline for generating DRInQ task datapoints.}
    \label{fig:detail}
\end{figure*}

A fundamental aspect of human communication is the ability to infer meaning  that goes beyond what is explicitly stated. Speakers routinely rely on shared common sense, social knowledge, and cooperative assumptions to encode and decipher these ``unspoken'' messages \cite{Grice1989-GRISIT, IndirectSpeechActs}. These inferences are typically predictable and reliable, allowing communication to remain efficient even in unfamiliar or unexpected situations~\cite{levinson_presumptive_2000}.



As conversational agents and AI assistants are increasingly deployed in everyday settings, they are expected to handle communication styles that extends beyond mere information delivery~\cite{srikanth-etal-2024-pregnant}. While pretrained large language models (LLMs) acquire substantial social commonsense understanding from training data alone~\cite{talmor-etal-2019-commonsenseqa}, they remain less competent at utilizing that information to make necessary inferences in veiled language use~\cite{setlur-chatbot2022, miehling-etal-2024-language}. One such phenomenon,
\textit{conversational implicature}, arises from how context reshapes meaning of an otherwise ordinary phrase, making it difficult to formalize and scale into an annotated resource~\cite{george2020conversational}.

We introduce \textbf{DRInQ} (Dataset for Recovering Implicature in Questions), a context-controlled benchmark for probing discriminative pragmatic reasoning in question utterances. Canonical examples such as the indirect request ``\textit{Can you pass the salt?}'' motivate our focus on questions, since interrogative forms routinely serve functions beyond literal inquiry~\cite{IndirectSpeechActs}. As illustrated in Fig~\ref{fig:intro}, circumstance alone can shift the impression listeners would draw from the same utterance, making such context-driven variation difficult to capture at scale. To address this challenge, we compose multiple contextual framings for each base question using a semi-automated, structured generation pipeline that supports scalable, context-sensitive data construction. This formulation enables a controlled study of how contextual variation reshapes implicature. We evaluate with multiple-choice contrasts as a diagnostic probe: they offer minimal pairs, efficient, low-variance scoring, and clear attribution, complementing recent generation-based measures of pragmatic competence.

Evaluating 12 state-of-the-art language models on \textbf{DRInQ}, we find that while models perform well overall, their inferences diverge systematically from human judgments. In particular, models frequently select interpretations that humans judge as overly strong or insufficiently supported by the context. On a challenging human-verified subset, the best-performing model (GPT-4o) reaches 76\% accuracy, compared to a human annotator average of 89\%. 
We further observe that structured prompting improves alignment with human judgments for smaller models, narrowing the performance gap with larger models. A comparative writing study reveals complementary strengths in data generation: human authors tend to produce conservative, underspecified contexts, whereas LLMs generate more diverse scenarios but often over-specify the implied interpretation. Together, these results highlight both the value of controlled, context-sensitive benchmarks for studying conversational implicature and the persistent challenges LLMs face in calibrating pragmatic inferences.


\section{Related Works}
Computational pragmatics is theoretically rooted in Grice's Cooperative Principle, which characterizes communication as guided by shared assumptions of cooperation~\cite{Grice1989-GRISIT, krause-vossen-2024-gricean-maxims}. Text records of non-cooperative language (e.g., sarcasm, deception) rarely explain the inferential process for recovering intended meaning~\cite{shwartz2020reporting, sravanthi2025understand, BANOU2025100192}, thus models often struggle when interpretation depends on unspoken conversational norms rather than surface semantics \cite{ruis2023goldilockspragmaticunderstandingfinetuning, Chang2023LanguageMB, pietro2023pragprof, tao2024chatgptroleplaydatasetanalysis}. Across pragmatic benchmarks, a consistent finding is that LLMs exhibit a strong \textit{literalist bias}, defaulting to surface-level interpretations even when pragmatic alternatives are licensed \cite{tong-etal-2024-metaphor, saakyan-etal-2025-understanding}. Empirically, partial linguistic cues used as scaffolding help SOTA LLMs on difficult indirect-meaning tasks (e.g., discourse relations, common ground)\cite{miao-etal-2024-discursive, qiu-etal-2025-wavelength, wan-etal-2025-role}, motivating continued development of resources that systematically provide and evaluate such cues.

\paragraph{Pragmatic Benchmarks and Tasks}
To address the scarcity of annotated pragmatic data, prior work has introduced datasets targeting non-literal language. Early efforts adapted natural language inference (NLI) formulations to probe pragmatic understanding, including social commonsense \cite{sap-etal-2019-social, zellers-etal-2019-hellaswag}, implicature and presupposition \cite{jeretic-etal-2020-imppres, stowe-etal-2022-impli}, scalar inference \cite{schuster-etal-2020-harnessing}, and indirect answers to polar questions \cite{louis2020idjustbedunderstanding}. 

Subsequent datasets expanded both scope and size, though these resources still focus on phenomena that are relatively well-defined or tied to identifiable lexical cues. IMPRES \cite{jeretic-etal-2020-imppres} and GRICE \cite{zheng-etal-2021-grice} provide controlled tests of implicature and presupposition via linguistically guided generation, while FLUTE \cite{chakrabarty2022flute, kulkarni-etal-2024-report, park-etal-2025-fluid} evaluates figurative language understanding including sarcasm, metaphor, and idiom. Aggregated evaluations show that instruction-tuned models approach human performance on some benchmarks, but results vary substantially across phenomena~\cite{sravanthi-etal-2024-pub}.

\paragraph{Linguisically-guided data generation}
\noindent  Most existing dataset classifications rely on coarse labels (i.e., literal vs.\ non-literal) which oversimplify meaningful distinctions, while more comprehensive studies require costly expert involvement to source, annotate, and validate~\cite{hu2023finegrainedcomparisonpragmaticlanguage}. Given the cost and difficulty of manual annotation, many pragmatic benchmarks rely on (semi-)synthetic data generation. Prior work has explored linguistically-grounded rule-based generation~\cite{jeretic-etal-2020-imppres, zheng-etal-2021-grice}, pattern-based extraction~\cite{parrish-etal-2021-nope, yue2024swordsmanimp}, and human-AI collaboration \cite{chakrabarty2022flute}. While such approaches enable scale, surveys note persistent challenges in maintaining pragmatic diversity and realism particularly for implicature~\cite{george2020conversational, ma-etal-2025-pragmatics}.

Our work builds on this literature by focusing specifically on \textit{question utterances}, where the divergence between surface form and communicative intent is often widest \cite{Yusupujiang2023whqs}. Unlike prior benchmarks that emphasize categorical labels and isolated phenomena, DRInQ targets fine-grained variation in conversational implicature induced solely by contextual differences. By combining a linguistically grounded intent framework with human-verified, model-in-the-loop generation, we aim to support scalable evaluation while preserving the multi-faceted characteristic of real-world pragmatic inference.

\section{The DRInQ Task}

\begin{figure}[t]
    \centering
    \includegraphics[width=.9\linewidth]{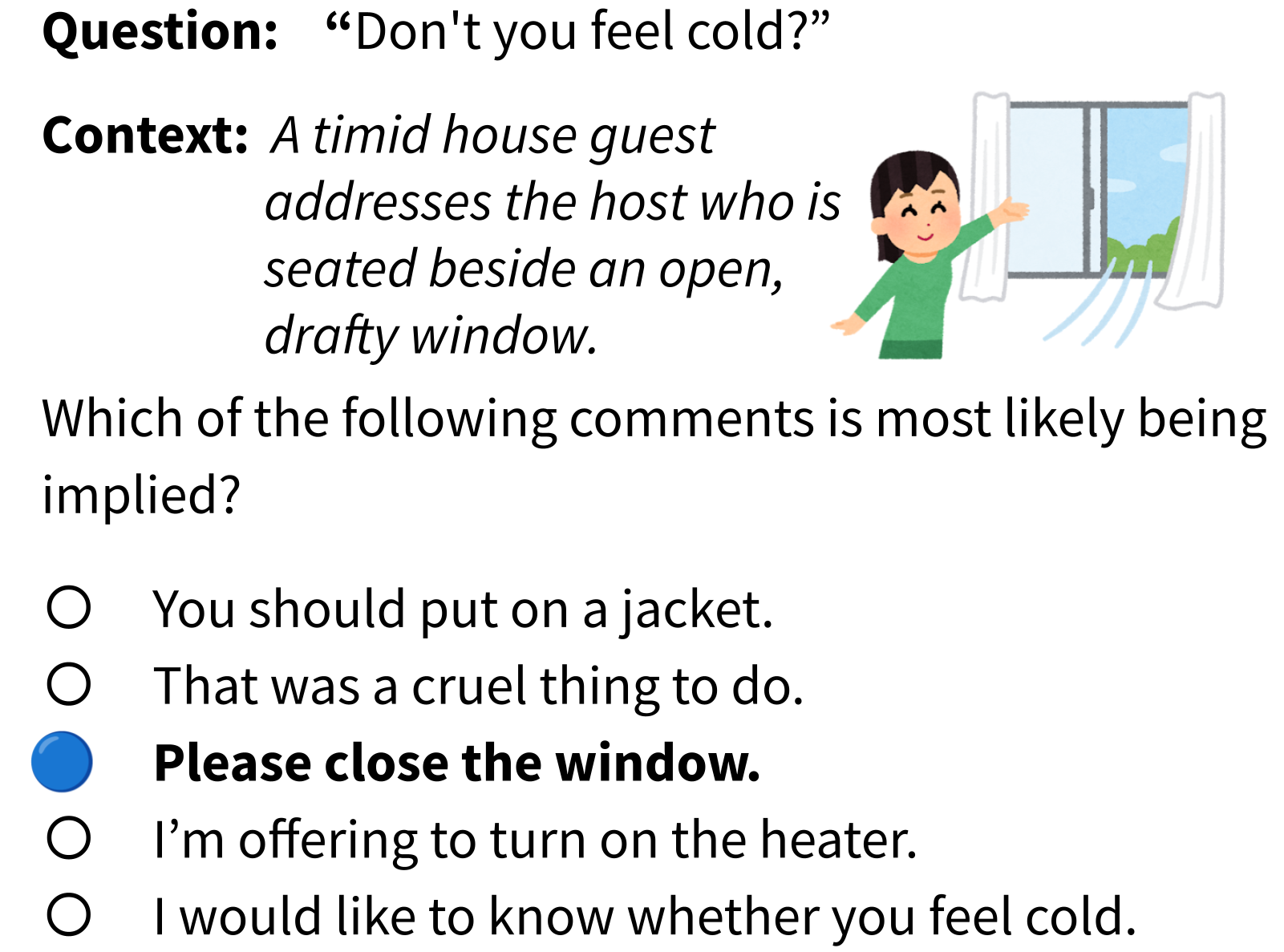}
    \caption{DRInQ task example. A single question utterance is paired with a contextual framing and multiple candidate interpretations.}
    \label{fig:taskeg}
\end{figure}

In natural conversation, implied meaning is often underspecified by utterance alone. A question like ``\textit{Don't you feel cold?}'' rarely functions as purely a literal inquiry; it may signal a rhetorical critique or indirectly prompt a specific action~\cite{IndirectSpeechActs} (see Fig.~\ref{fig:taskeg}). While a clarifying follow-up could resolve the ambiguity, such interventions are often less desirable, particularly when indirectness is motivated by politeness, face-saving, or efficiency~\cite{brown1987politeness}. 

\subsection{Evaluation Desiderata}
The reliance of conversational implicature on indirectness presents a distinct evaluation challenge for LLM inference. The difficulty is not in recognizing that an utterance \textit{can} convey implied meaning, but in determining \textit{which} contextual cues license a particular interpretation. While recent LLMs can readily track explicit state changes, they often struggle to separate pragmatically salient contextual information from incidental detail~\cite{shi2023large}. 

The failure to make this distinction amplifies the difficulty of designing and generating evaluation scenarios that probe this contrast. Although surface form wordings are easy to perturb, only a subset of contextual changes meaningfully affect interpretation \cite{ma-etal-2025-pragmatics}. For example, ``Can you pass the salt?'' remains an indirect request whether it is asked indoors or outdoors, while for other questions these same factors carry significant pragmatic weight.

To systematically control such interpretive shifts during dataset generation, we draw on the notion of \textit{speech acts} as a coarse-grained representation of a speaker's intended action. Following classic work in speech act theory~\cite{IndirectSpeechActs, searle1976classification, austin1975things}, we treat question utterances as capable of realizing distinct functions (e.g., requesting, criticizing, reassuring) depending on context, even when their surface form remains \textit{unchanged}. We use a structured inventory of speech act verbs (subsequently referred to as \textit{intent} labels) to guide the construction of contexts that license meaningfully different interpretations. For example, for the question in Figure~\ref{fig:intro}, scene 2 might have been produced with the intent to \textit{scold}, and scene 3 to \textit{confess}. Table~\ref{tab:speech-acts} presents a subset of the 23 intent labels we use; Appendix~\ref{sec:speechacts} provides additional details on the taxonomy and its role in the generation pipeline.

\begin{table}[t]
\small
\setlength{\tabcolsep}{3pt}
\renewcommand{\arraystretch}{1}
\centering
\begin{tabularx}{\columnwidth}{l >{\RaggedRight\arraybackslash}X >{\RaggedRight\arraybackslash}X}
\textbf{Category} & \textbf{Definition} & \textbf{Examples} \\ \midrule
Directive   & Commits the listener to some future task. & \textit{prohibit; request; seek information} \\
Assertive  & \makecell[tl]{Conveys some\\information.} & \textit{predict; rhetorical; report; conclude} \\
Commissive & Commits the speaker to some future task. & \textit{\makecell[tl]{promise; warn;\\ invite; offer}} \\
Expressive & Expresses the speaker's emotion. & \textit{thank; complain; apologize; insult} \\
\end{tabularx}
\caption{Intent labels (derived from speech act categories) used in the DRInQ generation pipeline}
\label{tab:speech-acts}
\end{table}

\subsection{Task Formulation}
\label{sec:form}
To translate the desiderata outlined above into a concrete evaluation setting, we formalize a multi-choice task that tests a model's ability to recover a speaker's intended meaning from a context. Each instance consists of a question utterance, a contextual framing, and a set of candidate interpretations, exactly one of which is licensed by the given context. 
Formally, each instance is represented as a tuple $(Q, C, \mathcal{A})$, where:
\begin{itemize}
\setlength{\itemsep}{0pt} 
    \item $Q$ is a \textit{question utterance}, i.e., the surface form of the query.
    \item $C$ is the \textit{context}, describing relevant situational and social factors relevant to the interpretation. 
    \item $\mathcal{A} = \{A_i\}_{i=1}^5$ is a set of \textit{implied comments} or candidate interpretations, with exactly one correct answer.
\end{itemize}
The remaining options are not arbitrary distractors: each corresponds to a plausible reading of $Q$ under a different contextual framing. As a result, success on this task cannot rely on lexical cues alone, but instead requires reasoning about how contextual factors modulate communicative intent. 

\noindent
The task design is guided by three principles:
\begin{enumerate}
\setlength{\itemsep}{0pt} 
    \item \textbf{Context sufficiency:} The context $C$ must provide enough information to uniquely determine the intended interpretation of $Q$.
    \item \textbf{Distractor plausibility:} Each incorrect $A_i$ must represent a reasonable interpretation of $Q$ in some alternative context.
    \item \textbf{Controlled variation:} Contextual manipulations should isolate pragmatically meaningful factors rather than incidental variation.
\end{enumerate}
\noindent
To satisfy these constraints, we adopt a minimal-contrast strategy common in linguistic analysis: the question $Q$ is held fixed while the context $C$ is systematically varied. This allows us to probe how different contextual factors license different implied meanings and to capture the range of interpretations a single utterance may convey.

These principles motivate a semi-automated generation pipeline (Fig.~\ref{fig:detail}). In the following, we detail each stage of this process, including base question curation, intent selection, context generation, task formatting, and human verification.
\begin{figure}
    \centering
    \includegraphics[width=1\linewidth]{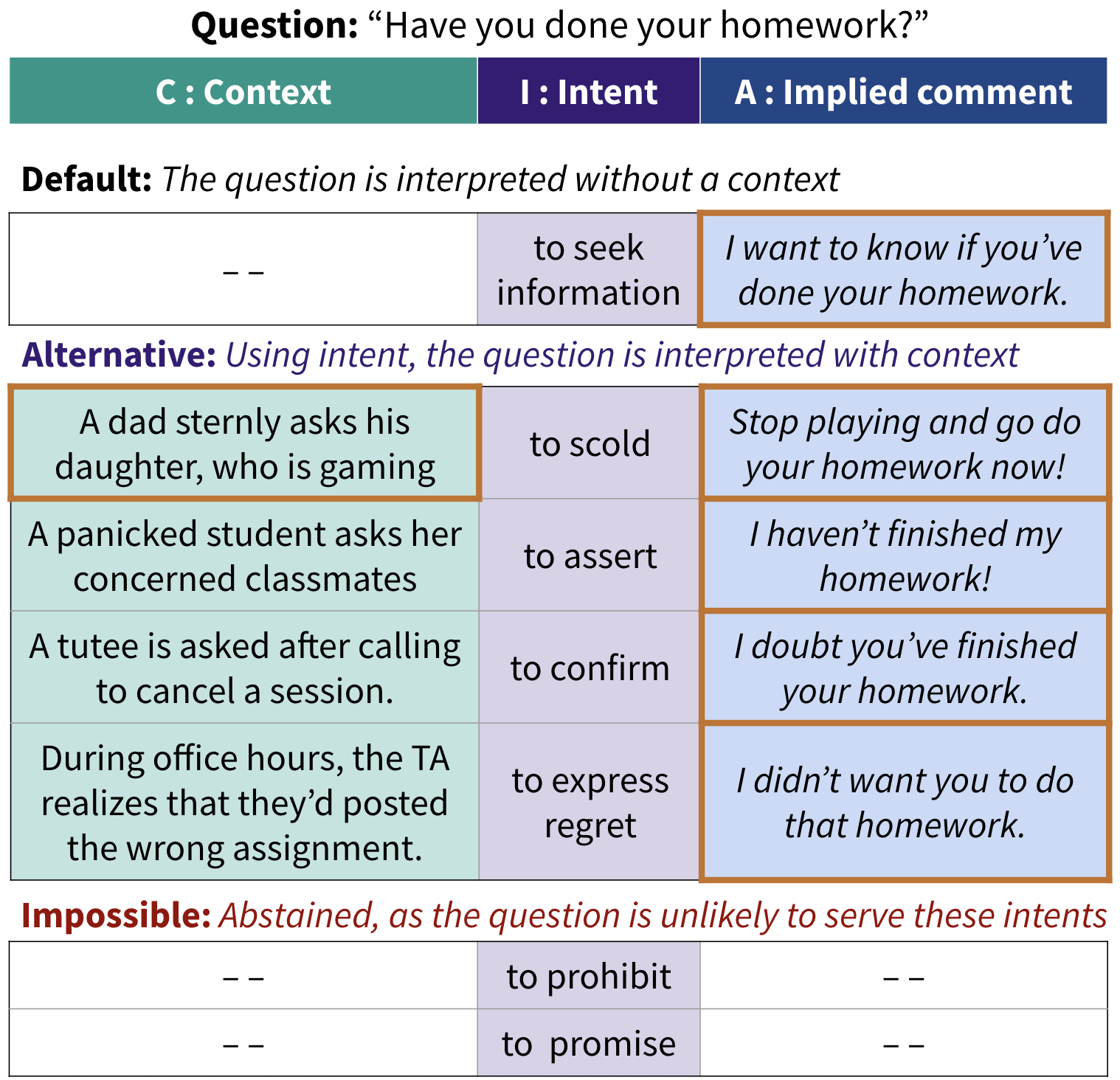}
    \caption{Example of text generated in Steps 3.3-3.6 of the pipeline on the base question ``Have you done your homework?'' The cells outlined in orange indicate the (\textit{C, A}) task components.}
    \label{fig:pipelinedetail}
\end{figure}
\subsection{Base Question Curation}
Most existing question datasets are QA- or task-oriented rather than casual conversation. As ensure natural, colloquial surface forms, we hand-selected 30 seed questions from everyday settings. We then used GPT-4o \cite{openai2024gpt4ocard} to expand these to 300 base questions (the fixed surface forms $Q$), constraining generation to an informal register, a mix of wh- or polar questions, and non–knowledge-heavy topics. A brief manual pass removed unnatural outputs and duplicated paraphrases. For each $Q$, we elicit a context-independent baseline interpretation with a default intent $I_d$ and default implied comment $A_d$, which serve as references when varying context.

\subsection{Intent Ranking and Selection}
For each question $Q$, we rank the remaining 22 intent labels by cosine distance between their semantic embeddings~\cite{reimers-2019-sentence-bert} to those of the default implied comment $A_d$. These ranked intents are used to seed the generation of novel contexts that license alternative interpretations (Figure~\ref{fig:intdist}). 

\subsection{Context Generation}
For each selected pair $(Q, I)$, we generate three context-interpretation pairs $\{(C_j, A_j)\}_{j=1}^3$. GPT-4o is provided with in-context examples sampled from a manually curated pool and is prompted to construct a scenario in which the speaker uses $Q$ to indirectly realize the target intent $I$. To maintain quality, the model is instructed to abstain from answering if a specific $Q \times I$ combination is pragmatically implausible. We retain only the ten highest-ranked $(Q, I)$ for which valid $(C, A)$ pairs are produced.
To ensure non-trivial implicature, we further filter out cases where the contextualized interpretation $A_g$ is semantically too similar to the default interpretation $A_d$. The final dataset contains 300 questions, each associated with at least five distinct intents and three context-interpretation variants per intent. Example few-shot example prompts are provided in Appendix~\ref{sec:appendix}.


\begin{figure}[!t]
    \centering
    \includegraphics[width=\linewidth]{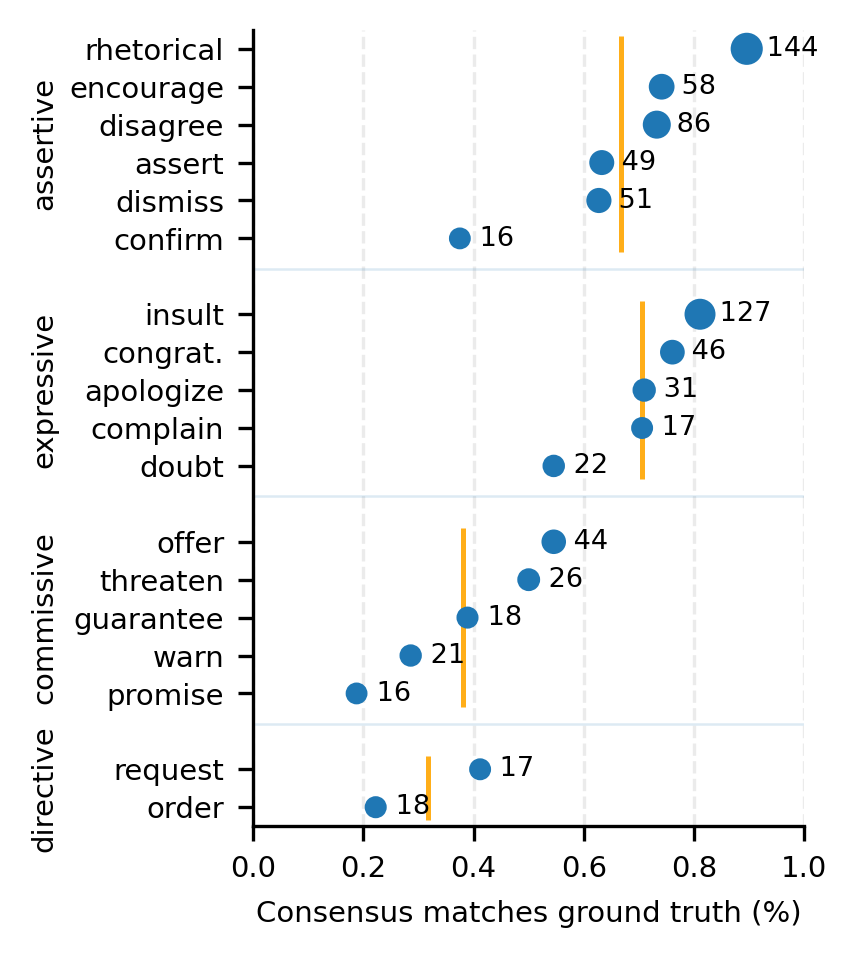}
    \caption{Proportion of question-context pairs $(Q,C)$ for which the human consensus $A_c$ matched the originally-generated comment $A_g$ by intent category. Dot size and numeric labels indicate the number of datapoints per intent. Vertical lines denote the category-wise mean.}
    \label{fig:goldcon}
\end{figure}

\subsection{Translating into Task Format}
We next map each generated context $C$ into the multiple-choice task format defined in Section \ref{sec:form}. Each instance is represented as a tuple $(Q, C, \mathcal{A})$, where
$\mathcal{A}
= \{ A^{k}_g, A_d \} \;\cup\; \{ A^{j}_g \mid j \neq k \}$.
Here, $A^{k}_g$ denotes the originally-generated implied comment conditioned on question $Q$ and target intent $I^{k}$, $A_d$ denotes the default implied comment for $Q$, and $A^{j}_g$ denotes implied comments corresponding to alternative intents $I^{j} \neq I^{k}$.
Superscripts index the conditioning intent, while subscripts distinguish default ($d$) from originally-generated, context-licensed ($g$) interpretations.

\subsection{Human Verification and Consensus}
To validate and measure human agreement on the dataset, we present the generated instances to 62 pre-screened annotators recruited via Prolific. We retain only datapoints for which at least 4 out of 5 annotators agree on the same implied comment, yielding 819 instances.

From the validated pool, we further identify a subset of 400 challenging instances for benchmarking. These tasks are characterized either by lower annotator-model agreement or by disagreement among annotators relative to the originally generated label, yielding a high-quality evaluation set for testing model performance on difficult pragmatic reasoning.

Under standard sampling, across this subset, agreement between the originally generated implied comment $A_g$ and the human consensus $A_c$ is 81\%. However, the final validated dataset is intentionally skewed towards more challenging instances, resulting in an overall agreement of 67\%. Within this set, agreement drops to 27\% on the hard baseline subset, but reaches 94\% on the remaining instances.   

Figure~\ref{fig:goldcon} summarizes agreement stratified by intent labels. Agreement is highest for questions functioning as rhetorical commentary (assertive) or as insults (expressive), where implied meaning is typically overt, approaching 90\%. In contrast, agreement is lower for commissive and directive uses, which require more nuanced reasoning about indirect offers or requests.

\section{Evaluation on Model Inference}

We evaluate LLM performance on \textbf{DRInQ} to assess their ability to recover conversational implicature from context. We report baseline results across state-of-the-art models, analyze systematic error patterns relative to human judgments, and examine whether targeted prompting scaffolds improve pragmatic inference.

\subsection{Baseline comparison of SOTA models}

\begin{table}[!t]
\centering
\setlength{\tabcolsep}{5pt}
\renewcommand{\arraystretch}{1.02}

\begin{tabular}{l
                S[table-format=1.2(2)]
                S[table-format=1.2(2)]}
\toprule
\textbf{Model}
& \multicolumn{1}{c}{\textbf{Vanilla}}
& \multicolumn{1}{c}{\textbf{Explanation}} \\
\midrule
GPT-5-Nano          & 0.45(7) & 0.43(2) \\
GPT-5-Mini          & 0.56(5) & 0.62(6) \\
GPT-4o              & 0.62(3) & 0.63(3) \\
GPT-4.1             & 0.61(2) & 0.61(3) \\
GPT-5               & 0.62(1) & 0.63(1) \\
OpenAI-o3           & 0.67(2) & 0.67(3) \\
Claude-Sonnet-4.5   & 0.65(1) & 0.62(2) \\
Claude-Haiku-4.5    & 0.64(1) & 0.63(2) \\
Llama-3.3-70B       & 0.58(2) & 0.58(3) \\
Qwen2.5-72B         & 0.56(3) & 0.57(3) \\
DeepSeek-V3         & 0.60(3) & 0.61(1) \\
DeepSeek-R1         & 0.62(4) & 0.60(4) \\
\midrule
Human Avg           & 0.88(10) & \multicolumn{1}{c}{--} \\
\bottomrule
\end{tabular}

\caption{Hard subset (400 items) accuracy: mean \(\pm\) SD over 3 runs (randomly sampling 3/6 in-context examples), vanilla vs.\ explanation prompting. 95\% item-level bootstrap CIs and paired significance in Appendix \ref{evaldetails}}
\label{tab:benchmark}
\end{table}

We benchmark twelve SOTA models from five model families on the DRInQ task. under two prompting conditions: \verb|vanilla|: a standard few-shot prompt with three in-context examples sampled from six manually written instances, and \verb|explanation|: an enhanced prompt that explicitly requires models to produce a written justification before answering. We adopt explanation-based prompting rather than chain-of-thought, following evidence that explicit step-by-step reasoning can hinder performance on tasks requiring intuitive inference~\cite{yao2025sarcasm, liu2025mindstepbystep}. We omit zero-shot or one-shot settings, as preliminary experiments indicate that minimal prompting fails to convey task conditions.

We report the results\footnote{Accuracies are computed on a fixed test set of 400 instances; differences of 1--2\% fall within statistical uncertainty. Model names are abbreviated for readability; full model identifiers are listed in Appendix~\ref{tab:model-mapping}} in Table~\ref{tab:benchmark}. Compared to the human annotator average baseline of 89\%, most SOTA models cluster within a narrow performance range around 60\%. Within the closed models, GPT-4o, which was used to generate the data, does well, but is out-performed by GPT-o3, as well as the models from Anthropic. Of the open-source models, DeepSeek-V3 and Llama-3.3-70B-Instruct-Turbo performed best with the \verb|vanilla| and \verb|explanation| variations, respectively. Overall, the reasoning-oriented variants from both the GPT and DeepSeek families outperform their chat-oriented counterparts, suggesting that enforcing explicit deliberation might aid in reasoning about conversational implicature.
\newcolumntype{Y}{>{\raggedright\arraybackslash}X}
\newcolumntype{Z}{>{\raggedright\arraybackslash}X}

\begin{table}[!t]
\small
\setlength{\tabcolsep}{3pt}
\renewcommand{\arraystretch}{1.35}
\centering

\begin{tabularx}{\columnwidth}{
>{\raggedright\arraybackslash}p{1.45cm}
>{\raggedright\arraybackslash\hsize=1.0\hsize}X
>{\raggedright\arraybackslash\hsize=1.0\hsize}X
}
\toprule
\textbf{Error Type} & \textbf{Malintent Attribution} & \textbf{Over-fixation on Detail} \\
\midrule

\textbf{Question} &
Do you need me to carry some of your shopping? &
Could you please turn down the volume? \\

\textbf{Context} &
\textit{An older sibling confidently asks their younger sibling, who is struggling with a heavy load of groceries.} &
\textit{A parent who ignored the child’s earlier request for louder music asks with a sheepish expression.} \\
\textbf{Human Consensus} &
I will help you, don’t worry. &
Please lower the volume. \\
\textbf{LLM Choice} &
You clearly can’t handle that by yourself. &
I apologize for not considering your love for music. \\
\bottomrule
\end{tabularx}
\caption{Representative error cases for LLMs (gpt-o3 and claude-haiku-5.4), contrasted with human consensus judgments.}
\label{fig:llmerror}
\end{table}

Across the benchmark, we observe systematic differences in how human annotators and LLMs resolve implied meaning. 

\paragraph{Malintent attribution:} Human readers tend to adopt a more charitable interpretation of speaker intent and tone: in the absence of explicit cues, they are reluctant to attribute hostile or morally negative intentions. A recurring model error arises when LLMs overemphasize isolated negative details, selecting a harsher interpretation than one supported by the human consensus (Table~\ref{fig:llmerror}, left). In this example, models appear to infer condescension from the speaker's ``confident demeanor,'' whereas annotators infer cooperative intent given the sibling dynamic.

\paragraph{Over-fixation on detail:} A second error pattern concerns \textit{contextual sufficiency}. In some cases, the originally-generated implied comment is logically compatible with the context but requires additional assumptions that human listeners typically do not make. While LLMs correctly identify relevant semantic cues, they often miscalibrate the strength of the inference, endorsing interpretations that are possible but pragmatically over-committed. In the situation shown in Table~\ref{fig:llmerror} (right), the model places disproportionate weight on the parent's ``sheepish'' expression, attributing it to regret over the earlier decision, while annotators instead interpret it as social awkwardness in making a repeated request. The divergence reflects not a a failure of semantic reasoning, but a mismatch in estimating which implicatures are warranted by the available evidence.

\section{Improved Prompts for the task}
\paragraph{Setup} To address the failure modes identified in the previous section, we evaluate prompt-based interventions aimed at eliciting more human-like pragmatic inference. Prior work suggests that errors on judgments  that are intuitive for humans can be mitigated by constraining how models reason through the presuppositions and inference \cite{sravanthi2025understand, weston2023system2}. Building on these insights, we test four targeted prompts that either directly counter the observed error patterns or explicitly scaffold pragmatic reasoning.

\begin{figure}[!t]
    \centering
    \includegraphics[width=\linewidth]{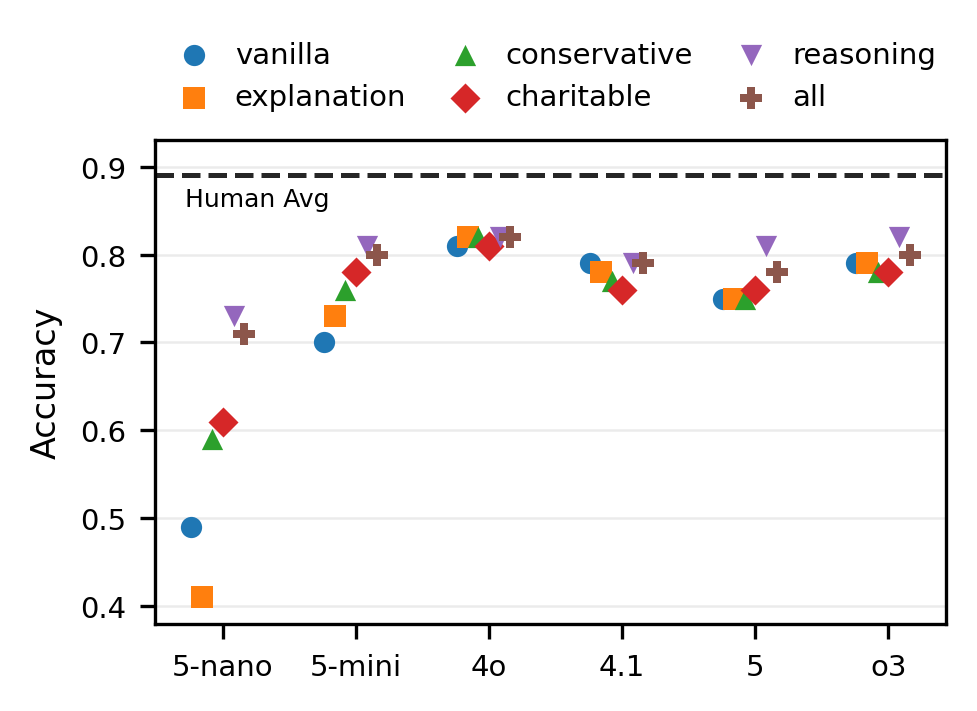}
    \caption{Benchmark performance of the four proposed structured prompting strategies on six GPT models. }
    \label{fig:thinkingprompt}
\end{figure}

The \textit{Conservativeness Constraint} prompt (\verb|conservative|) discourages over-commitment to merely compatible interpretations, while the \textit{Charitable Interpretation} prompt (\verb|charitable|) discourages unsupported attribution of hostile intent.
We additionally test a \textit{Pragmatic Reasoning Scaffold} prompt (\verb|reasoning|), inspired by prior work on constrained, task-aware prompting for intuitive linguistic judgments~\cite{yue2024swordsmanimp, yao2025sarcasm, lee2025pragmatic}, which models a cooperative listener's reasoning process, requiring the model to sequentially consider the surface question, contextual evidence, and inferred speaker intent before selecting an interpretation. The \textit{Combined Scaffold} (\verb|all|) condition combines all three. The detailed differences in the final prompts are available in Appendix \ref{app:evalprompt}.

\paragraph{Results} As shown in Figure~\ref{fig:thinkingprompt}, prompt-based interventions yield consistent improvements for smaller-capacity models, while having little to no effect on larger models; On the full annotated dataset, \texttt{gpt-5-nano} improves from 41\% to 73\% accuracy, and \texttt{gpt-5-mini} improves from 71\% to 81\%, narrowing the gap to the performance ceiling achieved by \texttt{gpt-4o} (82\%). 
Among the targeted strategies, \verb|conservative| and \verb|charitable| produce comparable gains, while the \verb|reasoning| prompt yields the largest performance increase. This pattern suggests that explicitly scaffolding pragmatic reasoning is especially beneficial under limited model capacity. Nevertheless, across all prompting conditions, performance remains below the human annotator average, indicating that prompt-based interventions mitigate but do not fully overcome capacity-related limitations in pragmatic inference.

\section{Comparing LLM and human writing}

\subsection{Human Baseline and Generation Study}

To assess the comparative quality of our machine-generated data, we conducted a human writing study on Prolific. The study mirrors the second stage of our generation pipeline: given question-intent pair $(Q, I)$, annotators were asked to write a contextual framing and a corresponding implied comment $(C, A_g)$. 
Annotators were screened using a qualification task assessing their ability to meet criteria of novelty, faithfulness, and sensibility. From our qualified pool, 16 annotators contributed writing for 64 $(Q, I)$ pairs drawn from the human-consensus subset of the dataset, with each pair written by three annotators, yielding a human-written baseline of 192 context-implied comment pairs\footnote{32 questions $\times$ 2 intents $\times$ 3 annotations}. Full instructions are provided in the Appendix (Fig~\ref{fig:writinginstruct}). On average, annotators required 2 minutes and 4 seconds per datapoint (SD: 55s), underscoring the cognitive demands of devising pragmatically appropriate conversational contexts.

\begin{figure}[h!]
    \centering
    \includegraphics[width=\linewidth]{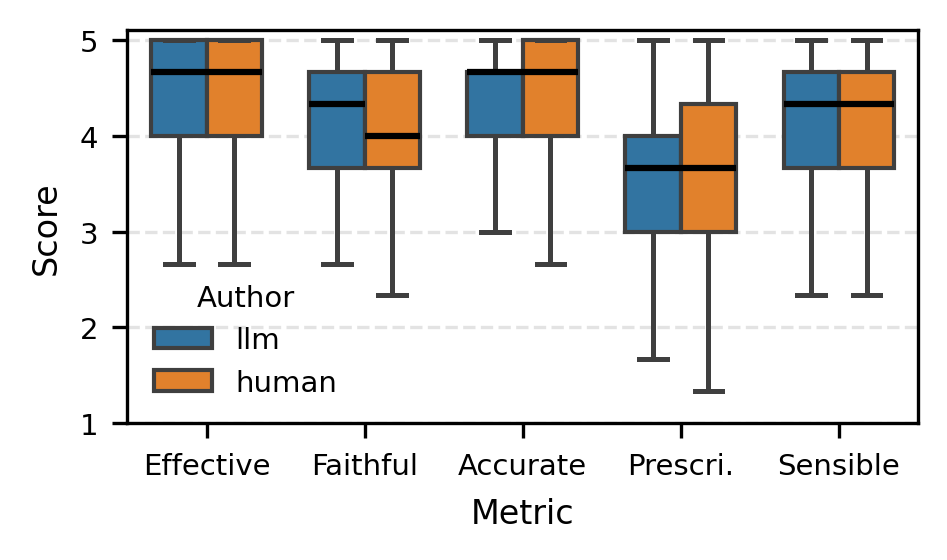}

        \caption{Distribution of ratings for LLM and human-authored context-interpretation pairs across five pragmatic quality metrics.}
    \label{fig:writingdist}
\end{figure}

\subsection{Context Quality Evaluation Methods}
\paragraph{Metrics}
To assess how well individual datapoints satisfy the desiderata outlined in Section~\ref{sec:form}, we conducted a separate human perception study assessing both human- and LLM-authored context-interpretation pairs $(C,A)$ on a 5-point Likert scale. Each metric operationalizes a distinct aspect of pragmatic quality targeted by our generation procedure: \emph{Effectiveness} (whether the context shifts interpretation away from the default baseline $A_d$), \emph{Faithfulness} (support for the specified intent $I$), \emph{Accuracy} (likelihood of inferring the target implied comment $A_g$), \emph{Non-prescriptiveness} (whether intent is implied rather than explicitly stated), and \emph{Sensibility} (internal coherence and commonsense plausibility). Because novelty is inherently comparative, we assess \emph{perceived novelty} using pairwise human judgments rather than absolute ratings. We define \textit{Novelty} as the degree to which a response is unexpected relative to plausible alternatives. Appendix~\ref{metricq} details the full metric definitions and annotation prompts, and Figure \ref{fig:writingdist} shows the distribution of scores across these dimensions for both human- and LLM-authored datapoints. 

For each of the 192 $(Q, I)$ pairs, we compare three human-written and three LLM-generated contexts. Annotators compare randomly sampled cross-group pairs and judge which context is more novel. The resulting nine pairwise judgments per $(Q, I)$ are aggregated to produce a group-level outcome. Overall, LLM-authored contexts are judged more novel in 37\% of cases, compared to 22\% for human-authored contexts, with no clear preference in 40\% of cases. 

\begin{table}[t]
\centering
\footnotesize
\setlength{\tabcolsep}{4pt}
\renewcommand{\arraystretch}{1.05}
\begin{tabularx}{\columnwidth}{l X}
\toprule
\multicolumn{2}{l}{\textbf{``Are we having dinner at home tonight?''} \textbf{\textit{(Insult)}}} \\
\midrule

\textbf{Human} &
A husband asks his wife after a failed home-cooked meal $\rightarrow$
\textit{The cooking is of a low standard.} \\
\midrule
\textbf{LLM} &
An adult child mockingly asks their parent after seeing a messy dining area $\rightarrow$
\textit{This place is a mess; you can’t manage the house.} \\
\bottomrule
\end{tabularx}
\caption{Human and LLM (GPT-4o) writings of context and interpretation for the same question and intent.}
\label{tab:writing}
\end{table}

\subsection{LLM vs.\ human-written situations}
Across evaluation dimensions, LLM-generated contexts largely overlap with human-written ones on \textit{Effectiveness}, \textit{Faithfulness}, and \textit{Sensibility}. Human writers score higher on \textit{Accuracy}, indicating closer alignment between their authored context and intended implied comment. Both groups exhibit similar central tendencies on \textit{Non-prescriptiveness}, though human-written contexts show higher variance, suggesting less consistent control over keeping internal motivations implicit. 
Overall, these results indicate a trade-off between contextual leverage and interpretive precision. LLM-generated contexts more reliably push  interpretations towards the specified intent, whereas human-written contexts more consistently support the intended comment. This pattern is illustrated in the human-authored example in Table~\ref{tab:writing}: although the intended interpretation is plausible, it depends on unstated assumptions about tone, relationship dynamics, and prior events. In the absence of these details, alternative readings (e.g., consoling a disappointed partner or expressing gratitude for the attempt) remain viable, reducing the perceived necessity of the intended implicature.

Novelty judgments further reveal complementary strengths. While many comparisons result in ties, LLM-generated contexts are more often judged novel, reflecting a greater willingness to explore non-canonical social configurations. In the LLM-authored example (Table~\ref{tab:writing}), a grown child mocks their parent -- an atypical arrangement that departs from familiar power dynamics. Human authors, by contrast, tend to default to more conventional role relations, producing scenarios that are less novel but more pragmatically stable.

Overall, these findings suggest that human authors prioritize interpretive plausibility grounded in shared social knowledge, whereas LLMs favor explicit contextual cues and exploratory configurations, yielding more overt but sometimes less intuitive implicatures.

\section{Conclusion}
We introduced \textbf{DRInQ}, a benchmark and semi-automated pipeline for constructing implicature-driven question-context-interpretation triples that enables controlled evaluation of pragmatic reasoning under contextual variation. By holding surface form constant while systematically manipulating context, \textbf{DRInQ} reveals interpretive distinctions that are difficult to observe with existing pragmatic benchmarks. 
Across evaluations, we find that many LLM errors reflect systematic over-commitment to inferred meaning, particularly in calibrating how much the context reasonably warrants beyond the literal question. For smaller models, structured prompting narrows the performance gap relative to the strongest benchmarks.
Our human-LLM writing study further highlights complementary strengths and weaknesses: human writers tend to produce more prototypical, low-risk contexts, while LLMs generate more diverse scenarios that over-specify implied meaning. Together, these results suggest that progress on pragmatic reasoning may require not only stronger models, but also evaluation frameworks that explicitly encode the underspecification inherent to everyday communication.

\section*{Limitations}
A growing body of recent work has established the inadequacies of multiple-choice task setups for measuring pragmatic reasoning in LLMs \cite{yerukola-etal-2024-pope, yu2026pragmaticmindmachinestracing}. While \textbf{DRInQ} aims to capture a broader range of context-dependent implicatures, it remains a discriminative evaluation with a fixed interpretation space, which may obscure alternative but reasonable pragmatic inferences. A further limitation concerns the relationship between our evaluation setup and the broader goal of conversational competence. Ultimately, pragmatic understanding should be assessed through a model's ability to generate appropriate responses in dialogue. Our multiple-choice formulation can of only approximate this ability, serving as a controlled proxy rather than a direct measure of generative implicature recovery. MCQ offers: (i) controlled minimal contrasts (ii) scalable evaluation and low-variance scoring and (iii) attribution (to identify \textit{which} alternative interpretation a model preferred). Thus, we view this task as a minimal-threshold diagnostic probe which complements generation-based evaluations of pragmatic competence.
In addition, our benchmark assumes that intent-conditioned interpretations generated during data construction constitute valid ground truth. While quality analysis and human agreement support this assumption in the majority of cases, we find that only 80.8\% of instances achieve our human consensus threshold of 4/5 annotators. This indicates that such interpretations cannot yet be treated as indisputable targets and underscores the need for future work that integrates uncertainty-aware or generative evaluation of pragmatic inference.

\section*{Ethics Statement} 

\paragraph{Human Subjects and Compensation} We recruited 78 participants via Prolific for dataset verification and writing tasks. All annotators were compensated at an hourly rate of at least \$12, in accordance with Prolific’s fair payment guidelines. We collected no personally identifiable information during this process, and all data were anonymized prior to analysis.
\paragraph{Cultural Limitations} We acknowledge that our dataset is English-centric and reflects Anglophone social norms~\cite{culturegen}. The annotation was conducted by our vetted Prolific participants, whose subjective judgments have been informed by their culture and language; we report the annotators' demographic distribution in Table \ref{tab:demographics}. Because conversational implicature is deeply culturally situated, this resource is unlikely to generalize to non-Western or low-context communicative settings. 
\paragraph{Downstream Harm} Improving LLM inference of user intent increases dual-use risks, including the amplification of \textit{dog whistles} or coded hate speech that evades moderation \cite{mendelsohn-etal-2023-dogwhistles}, as well as \textit{implicit profiling}, where models might infer sensitive user attributes from obstensively innocuous dialogue \cite{weidinger2021ethicalsocialrisksharm}. Despite the risk of misuse enabled by improvements in downstream models, we argue that the long-term benefits for pragmatic capabilities and defensive research outweigh the associated risks.
\paragraph{Environmental Impact} While our model-in-the-loop framework relies on API-inference rather than pretraining, we acknowledge the aggregate carbon footprint of large-scale querying \cite{strubell2019energy}. By releasing this benchmark as a static source, we aim to reduce the need for redundant high-volume data generation by future researchers.

\section*{Acknowledgments}
This research is supported in part by the Office of the Director of National Intelligence (ODNI), Intelligence Advanced Research Projects Activity (IARPA), via the HIATUS Program contract \#2022-22072200006, the Defense Advanced Research Projects Agency with award HR00112220046, and NSF IIS 2048211. We thank our annotators for their contribution, and would like to thank the collaborators at the INK research lab at USC for their constructive feedback on the work.
\bibliography{prag}

@inproceedings{ma-etal-2025-pragmatics,
    title = "Pragmatics in the Era of Large Language Models: A Survey on Datasets, Evaluation, Opportunities and Challenges",
    author = "Ma, Bolei  and
      Li, Yuting  and
      Zhou, Wei  and
      Gong, Ziwei  and
      Liu, Yang Janet  and
      Jasinskaja, Katja  and
      Friedrich, Annemarie  and
      Hirschberg, Julia  and
      Kreuter, Frauke  and
      Plank, Barbara",
    editor = "Che, Wanxiang  and
      Nabende, Joyce  and
      Shutova, Ekaterina  and
      Pilehvar, Mohammad Taher",
    booktitle = "Proceedings of the 63rd Annual Meeting of the Association for Computational Linguistics (Volume 1: Long Papers)",
    month = jul,
    year = "2025",
    address = "Vienna, Austria",
    publisher = "Association for Computational Linguistics",
    url = "https://aclanthology.org/2025.acl-long.425/",
    doi = "10.18653/v1/2025.acl-long.425",
    pages = "8679--8696",
    ISBN = "979-8-89176-251-0"
}

@inproceedings{strubell2019energy,
  title={Energy and Policy Considerations for Deep Learning in {NLP}},
  author={Strubell, Emma and Ganesh, Ananya and McCallum, Andrew},
  booktitle={Proceedings of the 57th Annual Meeting of the Association for Computational Linguistics},
  pages={3645--3650},
  year={2019},
  url={https://aclanthology.org/P19-1355}
}

@inproceedings{mendelsohn-etal-2023-dogwhistles,
    title = "From Dogwhistles to Bullhorns: Unveiling Coded Rhetoric with Language Models",
    author = "Mendelsohn, Julia  and
      Le Bras, Ronan  and
      Choi, Yejin  and
      Sap, Maarten",
    editor = "Rogers, Anna  and
      Boyd-Graber, Jordan  and
      Okazaki, Naoaki",
    booktitle = "Proceedings of the 61st Annual Meeting of the Association for Computational Linguistics (Volume 1: Long Papers)",
    month = jul,
    year = "2023",
    address = "Toronto, Canada",
    publisher = "Association for Computational Linguistics",
    url = "https://aclanthology.org/2023.acl-long.845/",
    doi = "10.18653/v1/2023.acl-long.845",
    pages = "15162--15180"
}

@article{culturegen,
  publtype={informal},
  author={Huihan Li and Liwei Jiang and Jena D. Huang and Hyunwoo Kim and Sebastin Santy and Taylor Sorensen and Bill Yuchen Lin and Nouha Dziri and Xiang Ren and Yejin Choi},
  title={CULTURE-GEN: Revealing Global Cultural Perception in Language Models through Natural Language Prompting},
  year={2024},
  cdate={1704067200000},
  journal={CoRR},
  volume={abs/2404.10199},
  url={https://doi.org/10.48550/arXiv.2404.10199}
}

@misc{weidinger2021ethicalsocialrisksharm,
      title={Ethical and social risks of harm from Language Models}, 
      author={Laura Weidinger and John Mellor and Maribeth Rauh and Conor Griffin and Jonathan Uesato and Po-Sen Huang and Myra Cheng and Mia Glaese and Borja Balle and Atoosa Kasirzadeh and Zac Kenton and Sasha Brown and Will Hawkins and Tom Stepleton and Courtney Biles and Abeba Birhane and Julia Haas and Laura Rimell and Lisa Anne Hendricks and William Isaac and Sean Legassick and Geoffrey Irving and Iason Gabriel},
      year={2021},
      eprint={2112.04359},
      archivePrefix={arXiv},
      primaryClass={cs.CL},
      url={https://arxiv.org/abs/2112.04359}, 
}

@book{austin1975things,
  title={How to Do Things with Words: Second Edition},
  author={Austin, J.L. and Urmson, J.O. and Sbis{\`a}, M.},
  isbn={9780674411524},
  lccn={75332044},
  series={Harvard paperback},
  url={https://books.google.com/books?id=V43VS07TGEMC},
  year={1975},
  publisher={Harvard University Press}
}

@misc{openai2024gpt4ocard,
      title={GPT-4o System Card}, 
      author={OpenAI},
      year={2024},
      eprint={2410.21276},
      archivePrefix={arXiv},
      primaryClass={cs.CL},
      url={https://arxiv.org/abs/2410.21276}, 
}

@book{levinson_presumptive_2000,
  title     = "Presumptive meanings",
  author    = "Levinson, Stephen C",
  publisher = "Bradford Books",
  series    = "Language, Speech, and Communication",
  month     =  apr,
  year      =  2000,
  address   = "Cambridge, MA",
  language  = "en"
}

@book{Grice1989-GRISIT,
	address = {Cambridge},
	author = {Herbert Paul Grice},
	editor = {},
	publisher = {Harvard University Press},
	title = {Studies in the Way of Words},
	year = {1989}
}

@article{george2020conversational,
  title={Conversational implicatures in English dialogue: Annotated dataset},
  author={George, Elizabeth Jasmi and Mamidi, Radhika},
  journal={Procedia Computer Science},
  volume={171},
  pages={2316--2323},
  year={2020},
  publisher={Elsevier}
}

@inproceedings{reimers-2019-sentence-bert,
    title = "Sentence-BERT: Sentence Embeddings using Siamese BERT-Networks",
    author = "Reimers, Nils and Gurevych, Iryna",
    booktitle = "Proceedings of the 2019 Conference on Empirical Methods in Natural Language Processing",
    month = "11",
    year = "2019",
    publisher = "Association for Computational Linguistics",
    url = "http://arxiv.org/abs/1908.10084",
}

@inproceedings{yerukola-etal-2024-pope,
    title = "Is the Pope Catholic? Yes, the Pope is Catholic. Generative Evaluation of Non-Literal Intent Resolution in {LLM}s",
    author = "Yerukola, Akhila  and
      Vaduguru, Saujas  and
      Fried, Daniel  and
      Sap, Maarten",
    editor = "Ku, Lun-Wei  and
      Martins, Andre  and
      Srikumar, Vivek",
    booktitle = "Proceedings of the 62nd Annual Meeting of the Association for Computational Linguistics (Volume 2: Short Papers)",
    month = aug,
    year = "2024",
    address = "Bangkok, Thailand",
    publisher = "Association for Computational Linguistics",
    url = "https://aclanthology.org/2024.acl-short.26/",
    doi = "10.18653/v1/2024.acl-short.26",
    pages = "265--275"
}

@article{Chang2023LanguageMB,
  title={Language Model Behavior: A Comprehensive Survey},
  author={Tyler A. Chang and Benjamin K. Bergen},
  journal={Computational Linguistics},
  year={2023},
  volume={50},
  pages={293-350},
  url={https://api.semanticscholar.org/CorpusID:257636789}
}

@inproceedings{tong-etal-2024-metaphor,
    title = "Metaphor Understanding Challenge Dataset for {LLM}s",
    author = "Tong, Xiaoyu  and
      Choenni, Rochelle  and
      Lewis, Martha  and
      Shutova, Ekaterina",
    editor = "Ku, Lun-Wei  and
      Martins, Andre  and
      Srikumar, Vivek",
    booktitle = "Proceedings of the 62nd Annual Meeting of the Association for Computational Linguistics (Volume 1: Long Papers)",
    month = aug,
    year = "2024",
    address = "Bangkok, Thailand",
    publisher = "Association for Computational Linguistics",
    url = "https://aclanthology.org/2024.acl-long.193/",
    doi = "10.18653/v1/2024.acl-long.193",
    pages = "3517--3536"
}

@inproceedings{saakyan-etal-2025-understanding,
    title = "Understanding Figurative Meaning through Explainable Visual Entailment",
    author = "Saakyan, Arkadiy  and
      Kulkarni, Shreyas  and
      Chakrabarty, Tuhin  and
      Muresan, Smaranda",
    editor = "Chiruzzo, Luis  and
      Ritter, Alan  and
      Wang, Lu",
    booktitle = "Proceedings of the 2025 Conference of the Nations of the Americas Chapter of the Association for Computational Linguistics: Human Language Technologies (Volume 1: Long Papers)",
    month = apr,
    year = "2025",
    address = "Albuquerque, New Mexico",
    publisher = "Association for Computational Linguistics",
    url = "https://aclanthology.org/2025.naacl-long.1/",
    doi = "10.18653/v1/2025.naacl-long.1",
    pages = "1--23",
    ISBN = "979-8-89176-189-6"
}

@inproceedings{kulkarni-etal-2024-report,
    title = "A Report on the {F}ig{L}ang 2024 Shared Task on Multimodal Figurative Language",
    author = "Kulkarni, Shreyas  and
      Saakyan, Arkadiy  and
      Chakrabarty, Tuhin  and
      Muresan, Smaranda",
    editor = "Ghosh, Debanjan  and
      Muresan, Smaranda  and
      Feldman, Anna  and
      Chakrabarty, Tuhin  and
      Liu, Emmy",
    booktitle = "Proceedings of the 4th Workshop on Figurative Language Processing (FigLang 2024)",
    month = jun,
    year = "2024",
    address = "Mexico City, Mexico (Hybrid)",
    publisher = "Association for Computational Linguistics",
    url = "https://aclanthology.org/2024.figlang-1.16/",
    doi = "10.18653/v1/2024.figlang-1.16",
    pages = "115--119"
}

@inproceedings{srikanth-etal-2024-pregnant,
    title = "Pregnant Questions: The Importance of Pragmatic Awareness in Maternal Health Question Answering",
    author = "Srikanth, Neha  and
      Sarkar, Rupak  and
      Mane, Heran  and
      Aparicio, Elizabeth  and
      Nguyen, Quynh  and
      Rudinger, Rachel  and
      Boyd-Graber, Jordan",
    editor = "Duh, Kevin  and
      Gomez, Helena  and
      Bethard, Steven",
    booktitle = "Proceedings of the 2024 Conference of the North American Chapter of the Association for Computational Linguistics: Human Language Technologies (Volume 1: Long Papers)",
    month = jun,
    year = "2024",
    address = "Mexico City, Mexico",
    publisher = "Association for Computational Linguistics",
    url = "https://aclanthology.org/2024.naacl-long.403/",
    doi = "10.18653/v1/2024.naacl-long.403",
    pages = "7253--7268"
}

@inproceedings{park-etal-2025-fluid,
    title = "{FLUID} {QA}: A Multilingual Benchmark for Figurative Language Usage in Dialogue across {E}nglish, {C}hinese, and {K}orean",
    author = "Park, Seoyoon  and
      Choi, Hyeji  and
      Kim, Minseon  and
      An, Subin  and
      Wang, Xiaonan  and
      Choi, Gyuri  and
      Kim, Hansaem",
    editor = "Christodoulopoulos, Christos  and
      Chakraborty, Tanmoy  and
      Rose, Carolyn  and
      Peng, Violet",
    booktitle = "Proceedings of the 2025 Conference on Empirical Methods in Natural Language Processing",
    month = nov,
    year = "2025",
    address = "Suzhou, China",
    publisher = "Association for Computational Linguistics",
    url = "https://aclanthology.org/2025.emnlp-main.1540/",
    doi = "10.18653/v1/2025.emnlp-main.1540",
    pages = "30268--30282",
    ISBN = "979-8-89176-332-6"
}

@article{BANOU2025100192,
title = {A systematic review of figurative language detection: Methods, challenges, and multilingual perspectives},
journal = {Natural Language Processing Journal},
volume = {13},
pages = {100192},
year = {2025},
issn = {2949-7191},
doi = {https://doi.org/10.1016/j.nlp.2025.100192},
url = {https://www.sciencedirect.com/science/article/pii/S2949719125000688},
author = {Zouheir Banou and Sanaa {El Filali} and El {Habib Benlahmar} and Fatima-Zahra Alaoui and Laila {El Jiani} and Hasnae Sakhi}
}

@inproceedings{zellers-etal-2019-hellaswag,
    title = "{H}ella{S}wag: Can a Machine Really Finish Your Sentence?",
    author = "Zellers, Rowan  and
      Holtzman, Ari  and
      Bisk, Yonatan  and
      Farhadi, Ali  and
      Choi, Yejin",
    editor = "Korhonen, Anna  and
      Traum, David  and
      M{\`a}rquez, Llu{\'i}s",
    booktitle = "Proceedings of the 57th Annual Meeting of the Association for Computational Linguistics",
    month = jul,
    year = "2019",
    address = "Florence, Italy",
    publisher = "Association for Computational Linguistics",
    url = "https://aclanthology.org/P19-1472/",
    doi = "10.18653/v1/P19-1472",
    pages = "4791--4800"
}

@book{brown1987politeness,
  title={Politeness: Some Universals in Language Usage},
  author={Brown, P. and Levinson, S.C.},
  isbn={9780521313551},
  lccn={lc86023255},
  series={Politeness: Some Universals in Language Usage},
  url={https://books.google.com/books?id=OG7W8yA2XjcC},
  year={1987},
  publisher={Cambridge University Press}
}

@inproceedings{Yusupujiang2023whqs,
author = {Yusupujiang, Zulipiye and Ginzburg, Jonathan},
year = {2023},
month = {01},
pages = {336-348},
title = {Unravelling Indirect Answers to Wh-Questions: Corpus Construction, Analysis, and Generation},
doi = {10.18653/v1/2023.sigdial-1.30}
}

@inproceedings{lee2025pragmatic,
  title={Pragmatic metacognitive prompting improves llm performance on sarcasm detection},
  author={Lee, Joshua and Fong, Wyatt and Le, Alexander and Shah, Sur and Han, Kevin and Zhu, Kevin},
  booktitle={Proceedings of the 1st Workshop on Computational Humor (CHum)},
  pages={63--70},
  year={2025}
}

@inproceedings{hu2023finegrainedcomparisonpragmaticlanguage,
    title = "A fine-grained comparison of pragmatic language understanding in humans and language models",
    author = "Hu, Jennifer  and
      Floyd, Sammy  and
      Jouravlev, Olessia  and
      Fedorenko, Evelina  and
      Gibson, Edward",
    editor = "Rogers, Anna  and
      Boyd-Graber, Jordan  and
      Okazaki, Naoaki",
    booktitle = "Proceedings of the 61st Annual Meeting of the Association for Computational Linguistics (Volume 1: Long Papers)",
    month = jul,
    year = "2023",
    address = "Toronto, Canada",
    publisher = "Association for Computational Linguistics",
    url = "https://aclanthology.org/2023.acl-long.230/",
    doi = "10.18653/v1/2023.acl-long.230",
    pages = "4194--4213"
}

@inproceedings{miao-etal-2024-discursive,
    title = "Discursive Socratic Questioning: Evaluating the Faithfulness of Language Models' Understanding of Discourse Relations",
    author = "Miao, Yisong  and
      Liu, Hongfu  and
      Lei, Wenqiang  and
      Chen, Nancy  and
      Kan, Min-Yen",
    editor = "Ku, Lun-Wei  and
      Martins, Andre  and
      Srikumar, Vivek",
    booktitle = "Proceedings of the 62nd Annual Meeting of the Association for Computational Linguistics (Volume 1: Long Papers)",
    month = aug,
    year = "2024",
    address = "Bangkok, Thailand",
    publisher = "Association for Computational Linguistics",
    url = "https://aclanthology.org/2024.acl-long.341/",
    doi = "10.18653/v1/2024.acl-long.341",
    pages = "6277--6295"
}

@inproceedings{talmor-etal-2019-commonsenseqa,
    title = "{C}ommonsense{QA}: A Question Answering Challenge Targeting Commonsense Knowledge",
    author = "Talmor, Alon  and
      Herzig, Jonathan  and
      Lourie, Nicholas  and
      Berant, Jonathan",
    editor = "Burstein, Jill  and
      Doran, Christy  and
      Solorio, Thamar",
    booktitle = "Proceedings of the 2019 Conference of the North {A}merican Chapter of the Association for Computational Linguistics: Human Language Technologies, Volume 1 (Long and Short Papers)",
    month = jun,
    year = "2019",
    address = "Minneapolis, Minnesota",
    publisher = "Association for Computational Linguistics",
    url = "https://aclanthology.org/N19-1421/",
    doi = "10.18653/v1/N19-1421",
    pages = "4149--4158"
}

@inproceedings{zheng-etal-2021-grice,
    title = "{GRICE}: A Grammar-based Dataset for Recovering Implicature and Conversational r{E}asoning",
    author = "Zheng, Zilong  and
      Qiu, Shuwen  and
      Fan, Lifeng  and
      Zhu, Yixin  and
      Zhu, Song-Chun",
    editor = "Zong, Chengqing  and
      Xia, Fei  and
      Li, Wenjie  and
      Navigli, Roberto",
    booktitle = "Findings of the Association for Computational Linguistics: ACL-IJCNLP 2021",
    month = aug,
    year = "2021",
    address = "Online",
    publisher = "Association for Computational Linguistics",
    url = "https://aclanthology.org/2021.findings-acl.182/",
    doi = "10.18653/v1/2021.findings-acl.182",
    pages = "2074--2085"
}

@inproceedings{stowe-etal-2022-impli,
    title = "{IMPLI}: Investigating {NLI} Models' Performance on Figurative Language",
    author = "Stowe, Kevin  and
      Utama, Prasetya  and
      Gurevych, Iryna",
    editor = "Muresan, Smaranda  and
      Nakov, Preslav  and
      Villavicencio, Aline",
    booktitle = "Proceedings of the 60th Annual Meeting of the Association for Computational Linguistics (Volume 1: Long Papers)",
    month = may,
    year = "2022",
    address = "Dublin, Ireland",
    publisher = "Association for Computational Linguistics",
    url = "https://aclanthology.org/2022.acl-long.369/",
    doi = "10.18653/v1/2022.acl-long.369",
    pages = "5375--5388"
}

@inproceedings{sravanthi-etal-2024-pub,
    title = "{PUB}: A Pragmatics Understanding Benchmark for Assessing {LLM}s' Pragmatics Capabilities",
    author = "Sravanthi, Settaluri  and Doshi, Meet  and Tankala, Pavan  and Murthy, Rudra  and Dabre, Raj  and Bhattacharyya, Pushpak",
    booktitle = "Findings of the Association for Computational Linguistics: ACL 2024",
    year = "2024",
    url = "https://aclanthology.org/2024.findings-acl.719"
}

@article{yao2025sarcasm,
  title={Is Sarcasm Detection a Step-by-Step Reasoning Process in Large Language Models?},
  author={Yao, Ben and Zhang, Yazhou and Li, Qiuchi and Qin, Jing},
  journal={Proceedings of the AAAI Conference on Artificial Intelligence},
  year={2025},
  url={https://ojs.aaai.org/index.php/AAAI/article/view/34756}
}

@article{liu2025mindstepbystep,
title = "Mind Your Step (by Step): Chain-of-Thought can Reduce Performance on Tasks where Thinking Makes Humans Worse",
author = "Ryan Liu and Jiayi Geng and Wu, \{Addison J.\} and Ilia Sucholutsky and Tania Lombrozo and Griffiths, \{Thomas L.\}",
note = "Publisher Copyright: {\textcopyright} 2025, by the authors.; 42nd International Conference on Machine Learning, ICML 2025 ; Conference date: 13-07-2025 Through 19-07-2025",
year = "2025",
language = "English (US)",
volume = "267",
pages = "38489--38517",
journal = "Proceedings of Machine Learning Research",
issn = "2640-3498",
publisher = "ML Research Press",
}

@inproceedings{chakrabarty2022flute,
    title = "{FLUTE}: Figurative Language Understanding through Textual Explanations",
    author = "Chakrabarty, Tuhin  and
      Saakyan, Arkadiy  and
      Ghosh, Debanjan  and
      Muresan, Smaranda",
    editor = "Goldberg, Yoav  and
      Kozareva, Zornitsa  and
      Zhang, Yue",
    booktitle = "Proceedings of the 2022 Conference on Empirical Methods in Natural Language Processing",
    month = dec,
    year = "2022",
    address = "Abu Dhabi, United Arab Emirates",
    publisher = "Association for Computational Linguistics",
    url = "https://aclanthology.org/2022.emnlp-main.481/",
    doi = "10.18653/v1/2022.emnlp-main.481",
    pages = "7139--7159"
}

@inproceedings{yue2024swordsmanimp,
  title={Do Large Language Models Understand Conversational Implicature -- A Case Study with a Chinese Sitcom},
  author={Yue, Shisen and Song, Siyuan and Cheng, Xinyuan and Hu, Hai},
  booktitle={Proceedings of the 23rd Chinese National Conference on Computational Linguistics},
  year={2024},
  url={https://aclanthology.org/2024.ccl-1.98/}
}

@inproceedings{sravanthi2025understand,
    title = "Understand the Implication: Learning to Think for Pragmatic Understanding",
    author = "Sravanthi, Settaluri Lakshmi  and
      Maharaj, Kishan  and
      Gunnu, Sravani  and
      Mishra, Abhijit  and
      Bhattacharyya, Pushpak",
    editor = "Che, Wanxiang  and
      Nabende, Joyce  and
      Shutova, Ekaterina  and
      Pilehvar, Mohammad Taher",
    booktitle = "Findings of the Association for Computational Linguistics: ACL 2025",
    month = jul,
    year = "2025",
    address = "Vienna, Austria",
    publisher = "Association for Computational Linguistics",
    url = "https://aclanthology.org/2025.findings-acl.1218/",
    doi = "10.18653/v1/2025.findings-acl.1218",
    pages = "23778--23790",
    ISBN = "979-8-89176-256-5"
}

@inproceedings{ruis2023goldilockspragmaticunderstandingfinetuning,
  title={The Goldilocks of Pragmatic Understanding: Fine-Tuning Strategy Matters for Implicature Resolution by LLMs},
  author={Laura Ruis and Akbir Khan and Stella Biderman and Sara Hooker and Tim Rocktaschel and Edward Grefenstette},
  year={2022},
  url={https://api.semanticscholar.org/CorpusID:253157310}
}

@inproceedings{shwartz2020reporting,
  title={Do Neural Language Models Overcome Reporting Bias?},
  author={Shwartz, Vered and Choi, Yejin},
  booktitle={Proceedings of the 28th International Conference on Computational Linguistics (COLING)},
  year={2020},
  url={https://aclanthology.org/2020.coling-main.605/}
}

@inproceedings{louis2020idjustbedunderstanding,
  title={“I’D Rather Just Go to Bed”: Understanding Indirect Answers},
  author={Annie Louis and Dan Roth and Filip Radlinski},
  booktitle={Conference on Empirical Methods in Natural Language Processing},
  year={2020},
  url={https://api.semanticscholar.org/CorpusID:222177178}
}

@inproceedings{jeretic-etal-2020-imppres,
    title = "Are Natural Language Inference Models {IMPPRES}sive? Learning {IMP}licature and {PRES}upposition",
    author = "Jeretic, Paloma  and Warstadt, Alex  and Bhooshan, Suvrat  and Williams, Adina",
    booktitle = "Proceedings of the 58th Annual Meeting of the Association for Computational Linguistics",
    year = "2020",
    url = "https://aclanthology.org/2020.acl-main.768/"
}

@inproceedings{schuster-etal-2020-harnessing,
    title = "Harnessing the linguistic signal to predict scalar inferences",
    author = "Schuster, Sebastian  and Chen, Yuxing  and Degen, Judith",
    booktitle = "Proceedings of the 58th Annual Meeting of the Association for Computational Linguistics",
    year = "2020",
    url = "https://aclanthology.org/2020.acl-main.479"
}

@inproceedings{parrish-etal-2021-nope,
    title = "{NOPE}: A Corpus of Naturally-Occurring Presuppositions in {E}nglish",
    author = "Parrish, Alicia  and Schuster, Sebastian  and Warstadt, Alex  and Agha, Omar  and Lee, Soo-Hwan  and Zhao, Zhuoye  and Bowman, Samuel R.  and Linzen, Tal",
    booktitle = "Proceedings of the 25th Conference on Computational Natural Language Learning",
    year = "2021",
    url = "https://aclanthology.org/2021.conll-1.28"
}

@inproceedings{tao2024chatgptroleplaydatasetanalysis,
  title={ChatGPT Role-play Dataset: Analysis of User Motives and Model Naturalness},
  author={Yufei Tao and Ameeta Agrawal and Judit Dombi and Tetyana Sydorenko and Jung In Lee},
  booktitle={International Conference on Language Resources and Evaluation},
  year={2024},
  url={https://api.semanticscholar.org/CorpusID:268723733}
}

@article{pietro2023pragprof,
author = {Barattieri di San Pietro, Chiara and Frau, Federico and Mangiaterra, Veronica and Bambini, Valentina},
year = {2023},
month = {08},
pages = {379-400},
title = {The pragmatic profile of ChatGPT: Assessing the communicative skills of a conversational agent},
volume = {XXXV},
journal = {Sistemi Intelligenti},
doi = {10.1422/108136}
}

@inproceedings{shi2023large,
  title={Large Language Models Can Be Easily Distracted by Irrelevant Context},
  author={Shi, Freda and Chen, Xinyun and Misra, Kanishka and Scales, Nathan and Dohan, David and Chi, Ed and Sch{\"a}rli, Nathanael and Zhou, Denny},
  booktitle={International Conference on Machine Learning (ICML)},
  year={2023},
  url={https://arxiv.org/abs/2302.00093}
}

@misc{yu2026pragmaticmindmachinestracing,
      title={The Pragmatic Mind of Machines: Tracing the Emergence of Pragmatic Competence in Large Language Models}, 
      author={Kefan Yu and Qingcheng Zeng and Weihao Xuan and Wanxin Li and Jingyi Wu and Rob Voigt},
      year={2026},
      eprint={2505.18497},
      archivePrefix={arXiv},
      primaryClass={cs.CL},
      url={https://arxiv.org/abs/2505.18497}, 
}

@inproceedings{miehling-etal-2024-language,
    title = "Language Models in Dialogue: Conversational Maxims for Human-{AI} Interactions",
    author = "Miehling, Erik  and
      Nagireddy, Manish  and
      Sattigeri, Prasanna  and
      Daly, Elizabeth M.  and
      Piorkowski, David  and
      Richards, John T.",
    editor = "Al-Onaizan, Yaser  and
      Bansal, Mohit  and
      Chen, Yun-Nung",
    booktitle = "Findings of the Association for Computational Linguistics: EMNLP 2024",
    month = nov,
    year = "2024",
    address = "Miami, Florida, USA",
    publisher = "Association for Computational Linguistics",
    url = "https://aclanthology.org/2024.findings-emnlp.843/",
    doi = "10.18653/v1/2024.findings-emnlp.843",
    pages = "14420--14437"
}

@inproceedings{setlur-chatbot2022,
author = {Setlur, Vidya and Tory, Melanie},
title = {How do you Converse with an Analytical Chatbot? Revisiting Gricean Maxims for Designing Analytical Conversational Behavior},
year = {2022},
isbn = {9781450391573},
publisher = {Association for Computing Machinery},
address = {New York, NY, USA},
url = {https://doi.org/10.1145/3491102.3501972},
doi = {10.1145/3491102.3501972},
booktitle = {Proceedings of the 2022 CHI Conference on Human Factors in Computing Systems},
articleno = {29},
numpages = {17},
location = {New Orleans, LA, USA},
series = {CHI '22}
}

@inproceedings{wan-etal-2025-role,
    title = "On the Role of Context for Discourse Relation Classification in Scientific Writing",
    author = "Wan, Stephen  and
      Liu, Wei  and
      Strube, Michael",
    editor = "Strube, Michael  and
      Braud, Chloe  and
      Hardmeier, Christian  and
      Li, Junyi Jessy  and
      Loaiciga, Sharid  and
      Zeldes, Amir  and
      Li, Chuyuan",
    booktitle = "Proceedings of the 6th Workshop on Computational Approaches to Discourse, Context and Document-Level Inferences (CODI 2025)",
    month = nov,
    year = "2025",
    address = "Suzhou, China",
    publisher = "Association for Computational Linguistics",
    url = "https://aclanthology.org/2025.codi-1.8/",
    doi = "10.18653/v1/2025.codi-1.8",
    pages = "96--106",
    ISBN = "979-8-89176-343-2"
}

@inproceedings{qiu-etal-2025-wavelength,
    title = "On the Same Wavelength? Evaluating Pragmatic Reasoning in Language Models across Broad Concepts",
    author = "Qiu, Linlu  and
      Zhang, Cedegao E.  and
      Tenenbaum, Joshua B.  and
      Kim, Yoon  and
      Levy, Roger P.",
    editor = "Christodoulopoulos, Christos  and
      Chakraborty, Tanmoy  and
      Rose, Carolyn  and
      Peng, Violet",
    booktitle = "Proceedings of the 2025 Conference on Empirical Methods in Natural Language Processing",
    month = nov,
    year = "2025",
    address = "Suzhou, China",
    publisher = "Association for Computational Linguistics",
    url = "https://aclanthology.org/2025.emnlp-main.1008/",
    doi = "10.18653/v1/2025.emnlp-main.1008",
    pages = "19913--19935",
    ISBN = "979-8-89176-332-6"
}

@inproceedings{krause-vossen-2024-gricean-maxims,
    title = "The {G}ricean Maxims in {NLP} - A Survey",
    author = "Krause, Lea  and
      Vossen, Piek T.J.M.",
    editor = "Mahamood, Saad  and
      Minh, Nguyen Le  and
      Ippolito, Daphne",
    booktitle = "Proceedings of the 17th International Natural Language Generation Conference",
    month = sep,
    year = "2024",
    address = "Tokyo, Japan",
    publisher = "Association for Computational Linguistics",
    url = "https://aclanthology.org/2024.inlg-main.39/",
    doi = "10.18653/v1/2024.inlg-main.39",
    pages = "470--485"
}

@inbook { IndirectSpeechActs,
      author = "John R. Searle",
      title = "Indirect Speech Acts",
      booktitle = "",
      year = "1975",
      publisher = "Brill",
      address = "Leiden, The Netherlands",
      isbn = "9789004368811",
      doi = "10.1163/9789004368811_004",
      pages=      "59 - 82",
      url = "https://brill.com/view/book/edcoll/9789004368811/BP000004.xml"
}

@article{searle1976classification,
  title={A classification of illocutionary acts},
  author={Searle, John R},
  journal={Language in society},
  volume={5},
  number={1},
  pages={1--23},
  year={1976},
  publisher={Cambridge University Press}
}

@inproceedings{sap-etal-2019-social,
    title = "Social {IQ}a: Commonsense Reasoning about Social Interactions",
    author = "Sap, Maarten  and Rashkin, Hannah  and Chen, Derek  and Le Bras, Ronan  and Choi, Yejin",
    booktitle = "Proceedings of the 2019 Conference on Empirical Methods in Natural Language Processing and the 9th International Joint Conference on Natural Language Processing (EMNLP-IJCNLP)",
    year = "2019",
    url = "https://aclanthology.org/D19-1454/"
}

@inproceedings{weston2023system2,
  title={System 2 Attention (is something you might need too)},
  author={Weston, Jason and Sukhbaatar, Sainbayar},
  booktitle={Advances in Neural Information Processing Systems},
  year={2023},
  url={https://arxiv.org/abs/2311.11829}
}
\clearpage
\clearpage
\section{Appendix}
\label{sec:appendix}

\subsection{Using Speech Acts as Intent Labels}
\label{sec:speechacts}
Drawing on Searle's speech acts \cite{searle1976classification, austin1975things}, we adopt a coarse-grained taxonomy of illocutionary acts as a way to characterize the functional roles a question utterance may serve in context. Although speech act theory is a rich and contested linguistic framework, we use it here as a practical abstraction to structure pragmatic variation during data generation.

When a speaker produces an utterance, its \textit{illocutionary force} corresponds to the action performed by speaking -- what they are \textit{doing} -- rather than the propositional content alone. For example, an utterance may function to insult, warn, or invite, even when its surface form is interrogative. Individual act verbs can be grouped into broader \textbf{illocutionary points} which reflect their its essential communicative purpose. 

\begin{table}[th]
\small
\setlength{\tabcolsep}{4pt}
\renewcommand{\arraystretch}{1.3}

\centering
\begin{tabularx}{\columnwidth}{l >{\RaggedRight\arraybackslash}X >{\RaggedRight\arraybackslash}X}
\textbf{Act Category} & \textbf{Definition} & \textbf{Examples} \\ \midrule
Directive   & \makecell[tl]{Commits the \\listener to some \\future action.} & \textit{prohibit; request; seek information; order; advise} \\
Assertive  & \makecell[tl]{Represents a state \\of affairs as true.} & \textit{predict; report; conclude; inform; make rhetorical comment} \\
Commissive & Commits the speaker to some future action. & \textit{promise; warn; invite; threaten; offer; guarantee} \\
Expressive & Expresses the speaker's psychological state or emotion. & \textit{thank; complain; apologize; insult; greet; support} \\
\end{tabularx}
\caption{Speech act categories used}
\label{tab:speech-acts}
\end{table}

We consider four illocutionary points, \textbf{Directives, Assertives, Commissives, and Expressives}, from which we aggregate 23 representative act verbs to support controlled variation in pragmatic interpretation. (Table~\ref{tab:speech-acts}). We omit the fifth category \textbf{Declarations} (brings about an institutional change in the world, e.g. \textit{"I resign"} or \textit{"You are fired"}), as such acts rarely present as a question.

\subsection{Prompt Templates for Generating context-interpretation $(C,I)$ pairs:}
\label{app:prompts}

\paragraph{Instructions:}

\begin{promptbox}
You are an insightful, creative expert in pragmatics. Often, people will say one thing, but mean something else.

You are given a "question" and "intent".

Imagine a scenario in which someone asks this "question," but with a deeper meaning because they had the goal to "intent". Your job is to generate a "context" that would make the "intent" true and obvious, then identify the "implied comment". 

"context" is a description of the situation in which the question is asked (e.g. the speaker's tone or expression, the speaker and listeners' relationship or power dynamic, social conventions, history of the conversation, past events, location etc.), which makes the "intended comment" apparent. "context" should not include interpretations/implications of the situation, nor the speaker's internal hopes, nor information about the future.

"Implied comment" is the message being implied by the "question," as a brief statement.

Abstain from generating (respond with an empty list) if the "question" could never be used to imply "implied comment". Otherwise, generate 3 sets of these answers, formatted as a list of dictionaries."
\end{promptbox}

\paragraph{In-context Examples}
\begin{promptbox}
Question 1: Are you going out?
Context: The speaker points at a full garbage can.  
Implied meaning: Please take out the trash on your way out.
Context: The speaker looks at the clock, which shows midnight. 
Implied meaning: Why are you leaving the house so late?
Context: The listener is wearing pajamas.  
Implied meaning: Why are you leaving the house in pajamas?

Question 2: Do you have a pencil?
Context: The speaker is in class without a pencil.  
Implied meaning: Please let me borrow your pencil.
Context: The listener needs to write something down.  
Implied meaning: Would you like to borrow my pencil?
Context: The listener is writing on a plastic bottle.  
Implied meaning: Why are you trying to write on plastic with a pencil?
\end{promptbox}

\subsection{Intent Selection for Context Generation}
\begin{figure*}[h]
    \centering
    \includegraphics[width=0.49\textwidth]{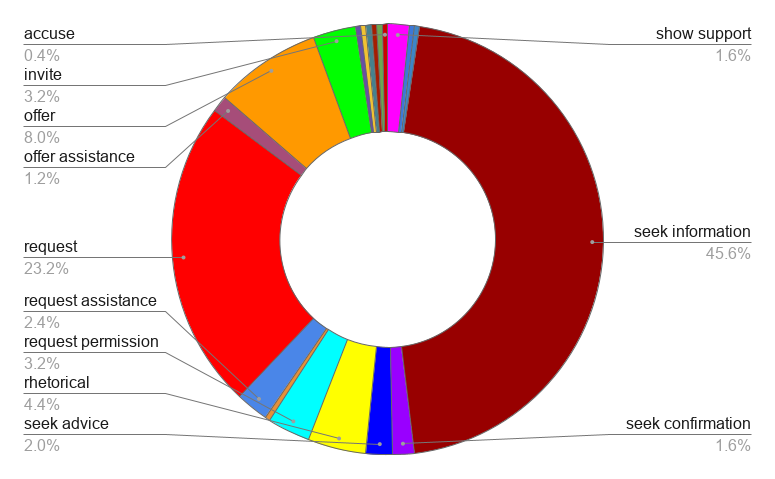}
    \includegraphics[width=0.49\textwidth]{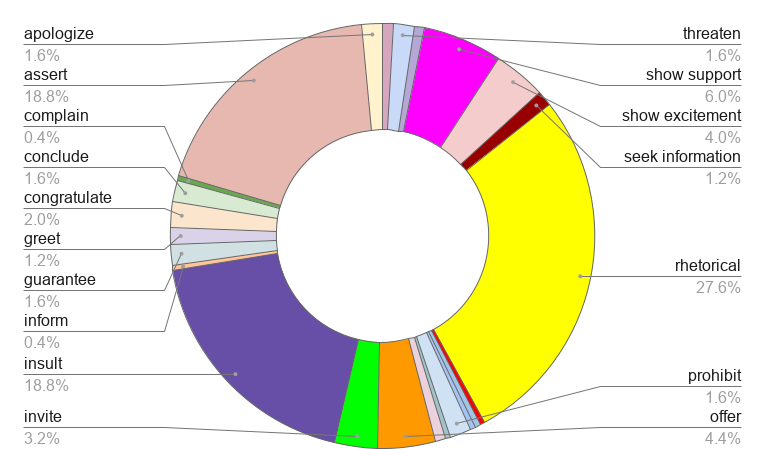}
    \caption{Distribution of intent labels in the dataset. 
The left panel shows default intents for base questions, while the right panel shows the most semantically dissimilar intents selected to seed context generation.}
    \label{fig:intdist}
\end{figure*}

To induce diverse pragmatic interpretations while holding the surface form of each question fixed, we seed context generation with intents that are semantically distant from the question’s default reading. Figure~\ref{fig:intdist} provides supporting statistics on the resulting intent distributions.
\subsection{Prompt Templates for Evaluation}
\label{app:evalprompt}
\paragraph{Core Prompt}
\begin{promptbox}
Instructions:
Based on the conversational context, choose the option from the given list that most accurately reflects the speaker's intended meaning when they ask the question.

## Output Format
Return your response as a JSON object in the following format:
```
{
  "selected_option": "<letter and full text of the selected option>",
  "explanation": "<concise explanation for why this option was selected>"
}
```
Replace `<letter and full text of the selected option>` with your choice, using the format provided in the list. The explanation should briefly justify your selection. Ensure the output strictly matches the specified JSON structure. If the output structure deviates, self-correct before returning your final answer.

\end{promptbox}

\paragraph{\textit{Pragmatic Reasoning} prompt}

\begin{promptbox}
Add instructions:
1. Identify the intent of the question as stated, without considering context.
2. Now incorporate context, reasoning whether and how the given context alters the perceived intent.
3. Select the most appropriate implied comment from the provided options, ensuring your choice reflects both the original question and context.
\end{promptbox}

\paragraph{\textit{Charitable Interpretation} prompt}
\begin{promptbox}
Add constraint: Be generous in your interpretation of the speaker's intent. Do not misattribute malicious intent where not supported by the context.
\end{promptbox}

\paragraph{\textit{Conservativeness Constraint} prompt}
\begin{promptbox}
Add constraint: Distinguish between details that are present and those that are pragmatically salient. Do not infer an implication from a detail unless it is clearly relevant to the act of asking the question.
\end{promptbox}

\subsection{Specifications of the Human Study}

\paragraph{Human Verification Agreement by Intent Label}

\begin{figure*}[h]
    \centering
    \includegraphics[width=0.9\textwidth]{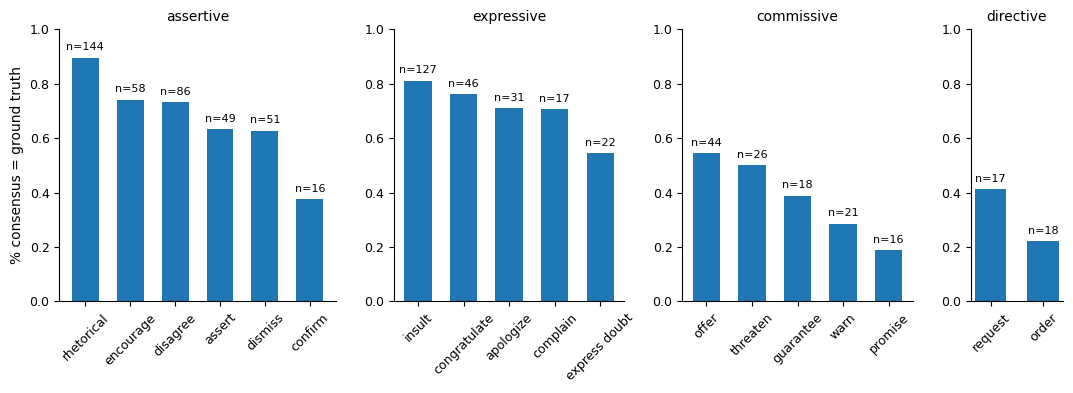}
    \caption{Ratio of datapoints for which the annotator consensus is the orignally-generated interpretation}
    \label{fig:intentbreakdown}
\end{figure*}

Figure~\ref{fig:intentbreakdown} reports agreement between the originally generated implied comment $A_g$ and the human consensus $A_c$, stratified by intent label.

\paragraph{Prolific Annotator Demographics}

\begin{table}[ht]
\centering
\footnotesize
\setlength{\tabcolsep}{0pt} 

\begin{tabularx}{\columnwidth}{
    X                         
    >{\raggedleft\arraybackslash}p{2.2em} 
    >{\raggedleft\arraybackslash}p{3.5em} 
    @{\hspace{6pt}}|@{\hspace{6pt}}        
    X                         
    >{\raggedleft\arraybackslash}p{2.2em} 
    >{\raggedleft\arraybackslash}p{3.5em} 
}
\toprule
\textbf{Category} & \textbf{n} & \textbf{(\%)} & \textbf{Category} &  \textbf{n} & \textbf{(\%)} \\
\midrule
\textbf{Sex} & & & \textbf{Student} & & \\
\hspace{2mm}Male & 31 & (50.0) & \hspace{2mm}No & 49 & (79.0) \\
\hspace{2mm}Female & 29 & (46.8) & \hspace{2mm}Yes & 3 & (4.8) \\
\hspace{2mm}DNS & 2 & (3.2) & \hspace{2mm}DNS & 10 & (16.2) \\
\midrule
\textbf{Language} & & & \textbf{Residence} & & \\
\hspace{2mm}English & 57 & (91.9) & \hspace{2mm}USA & 54 & (87.1) \\
\hspace{2mm}Other & 5 & (8.1) & \hspace{2mm}Other & 8 & (12.9) \\
\midrule
\textbf{Employment} & & & \textbf{Age} & & \\
\hspace{2mm}Full-Time & 28 & (45.2) & \hspace{2mm}Mean (SD) & \multicolumn{2}{r}{45.3 (13.5)} \\
\hspace{2mm}Part-Time & 9 & (14.5) & \hspace{2mm}Range & \multicolumn{2}{r}{20--77} \\
\hspace{2mm}Not paid & 10 & (16.1) & & & \\
\hspace{2mm}Other / DNS & 15 & (24.2) & & & \\
\bottomrule

\end{tabularx}
\caption{Aggregated demographics of screened study participants on Prolific ($N=62$). DNS indicates Data Not Shared.}
\label{tab:demographics}
\end{table}

\begin{figure*}[h]
    \centering
    \includegraphics[width=\textwidth]{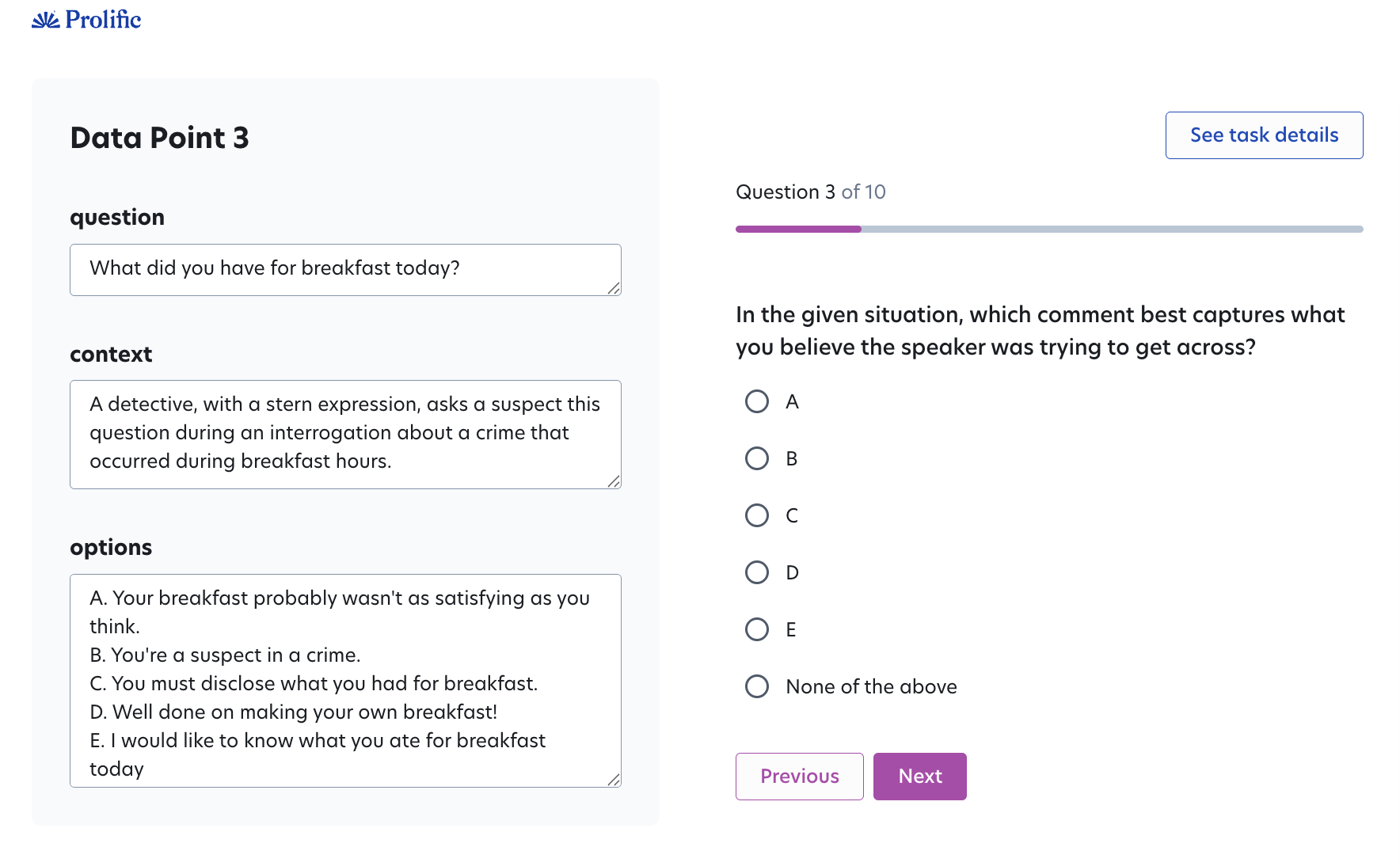}
    \caption{Interface for the DRInQ task on Prolific}
    \label{fig:taskhuman}
\end{figure*}

\begin{figure*}[h]
    \centering
    \includegraphics[width=0.45\textwidth]{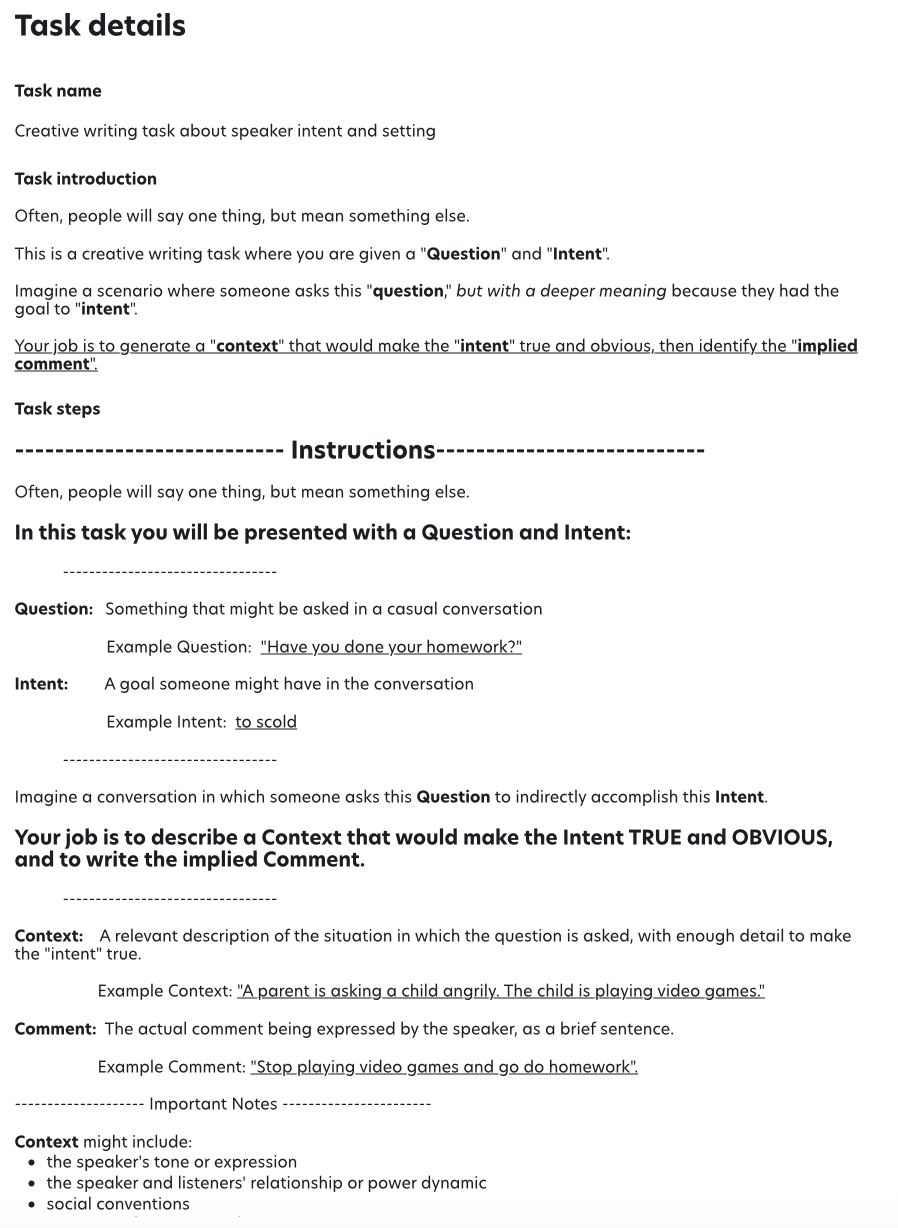}
    \includegraphics[width=0.45\textwidth]{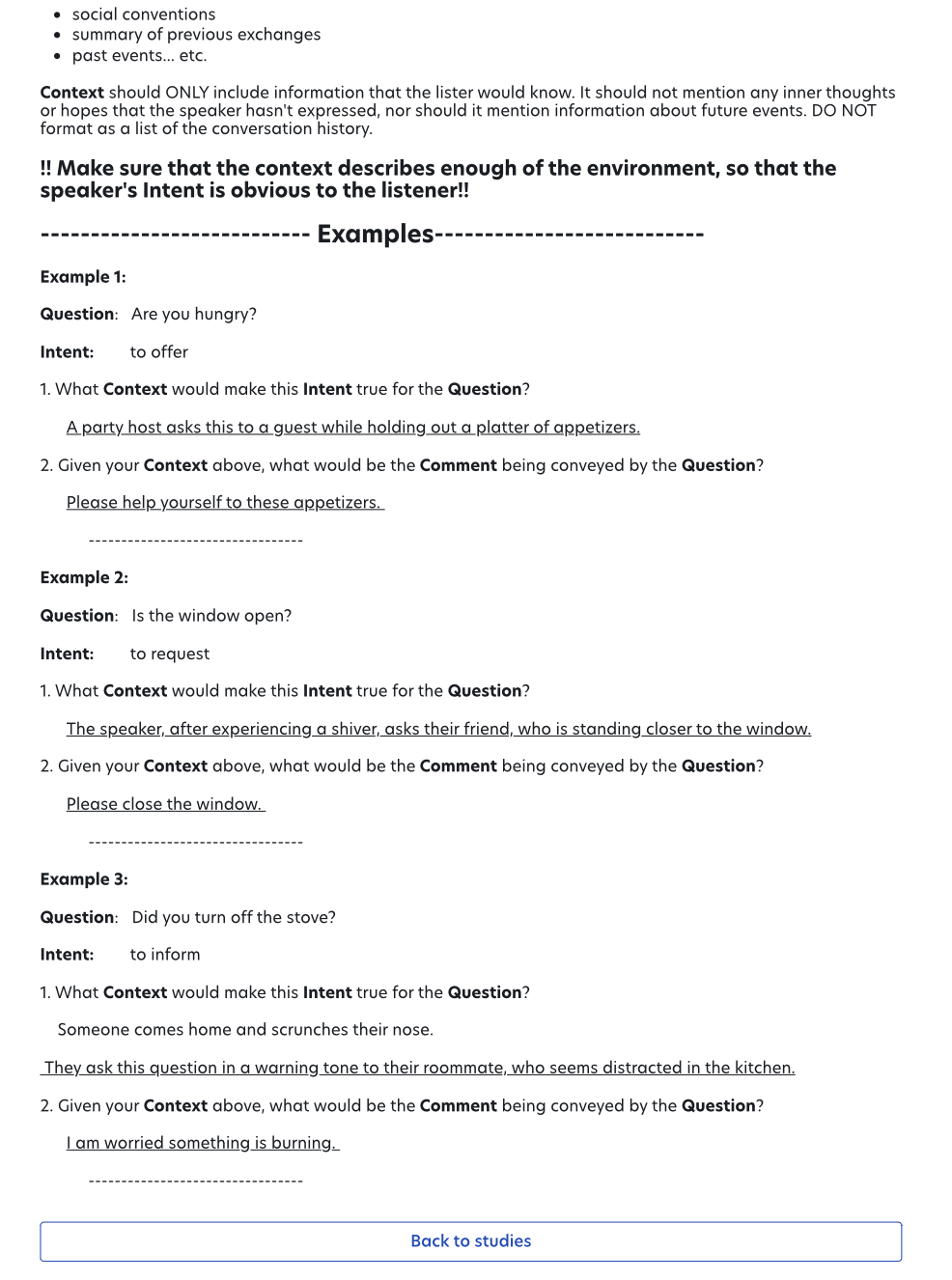}
    \caption{Instructions for the writing study on Prolific}
    \label{fig:writinginstruct}
\end{figure*}

\paragraph{Annotation Interfaces and Instructions}
Figure~\ref{fig:taskhuman} shows the instructions provided to annotators for the DRInQ data validation, and Figure~\ref{fig:writinginstruct} shows the instructions for the writing study.

\subsection{Model Identifiers and Evaluation Details}
\label{evaldetails}
\begin{table}[h]
\centering
\small
\caption{Mapping between abbreviated model names used in the main paper and their full model identifiers}
\setlength{\tabcolsep}{1pt}
\renewcommand{\arraystretch}{1.1}
\begin{tabular}{ll}
\toprule
\textbf{Model Label} & \textbf{Full  Identifier} \\
\midrule
GPT-5-Nano & OpenAI GPT-5 Nano \\
GPT-5-Mini & OpenAI GPT-5 Mini \\
GPT-5 & OpenAI GPT-5 \\
GPT-4o & OpenAI GPT-4o \\
GPT-4.1 & OpenAI GPT-4.1 \\
OpenAI-o3 & OpenAI o3 Reasoning Model \\
Claude-Haiku-4.5 & Anthropic Claude Haiku 4.5 \\
Claude-Sonnet-4.5 & Anthropic Claude Sonnet 4.5 \\
Llama-3.3-70B & meta-llama/Llama-3.3-70B-Instruct-Turbo \\
Qwen2.5-72B & Qwen/Qwen2.5-72B-Instruct-Turbo \\
DeepSeek-V3 & deepseek-ai/DeepSeek-V3 \\
DeepSeek-R1 & deepseek-ai/DeepSeek-R1 \\
\bottomrule
\end{tabular}

\label{tab:model-mapping}
\end{table}

\begin{table}[t]
\centering
\footnotesize
\setlength{\tabcolsep}{3pt}
\caption{Accuracy on the hard subset (400 items). Mean $\pm$ SD over 3 runs (randomly sampling 3 of 6 in-context examples); 95\% item-level bootstrap CIs in brackets.}
\label{tab:hard-subset-accuracy}
\begin{tabularx}{\columnwidth}{
>{\raggedright\arraybackslash}p{1.8cm}
>{\centering\arraybackslash\hsize=1\hsize}X
>{\centering\arraybackslash\hsize=1\hsize}X
}
\toprule
Model & Vanilla & Explanation \\
\midrule
gpt-5-nano & .45 $\pm$ .07 [.39, .50] & .43 $\pm$ .02 [.37, .49] \\
gpt-5-mini & .56 $\pm$ .05 [.51, .62] & .62 $\pm$ .06 [.57, .67] \\
gpt-4o & .62 $\pm$ .03 [.56, .67] & .63 $\pm$ .03 [.57, .68] \\
gpt-4.1 & .61 $\pm$ .02 [.55, .67] & .61 $\pm$ .03 [.56, .67] \\
gpt-5 & .62 $\pm$ .01 [.57, .68] & .63 $\pm$ .01 [.57, .69] \\
o3 & .67 $\pm$ .02 [.61, .71] & .67 $\pm$ .03 [.61, .72] \\
sonnet-4-5 & .65 $\pm$ .01 [.59, .70] & .62 $\pm$ .02 [.57, .67] \\
haiku-4-5 & .64 $\pm$ .00 [.59, .70] & .63 $\pm$ .02 [.58, .68] \\
llama-3.3 & .58 $\pm$ .02 [.52, .63] & .58 $\pm$ .03 [.53, .64] \\
qwen2.5-72B & .20 $\pm$ .34 [.18, .22] & .21 $\pm$ .36 [.19, .23] \\
deepseek-V3 & .60 $\pm$ .03 [.55, .66] & .61 $\pm$ .01 [.55, .66] \\
deepseek-R1 & .62 $\pm$ .04 [.56, .67] & .60 $\pm$ .04 [.55, .66] \\
Human Avg & .88 $\pm$ .10 [.87, .89]& \\
\bottomrule
\end{tabularx}
\end{table}

\subsection{Likert-Scale Statements for Dataset Quality Metrics}
\label{metricq}

Each context-interpretation $(C,A)$ pair was evaluated on a 5-point Likert scale using the following statements:

\begin{enumerate}[leftmargin=*, itemsep=6pt]
    \item \textbf{Effective}: The degree to which the provided context $C$ meaningfully alters the interpretation of the question relative to its context-free baseline $A_d$.
    
    \textit{I would interpret the question \emph{with} this context differently than the same question \emph{without} the context.}
    
    \item \textbf{Faithful}: The extent to which the context supports the specified communicative intent $I$, making the speaker’s goal inferable from the scenario.
    
    \textit{Based on the context, it would be clear to me that the speaker is using the question to accomplish this intent.}
    
    \item \textbf{Accurate}: The likelihood that a listener would infer the target implied comment $A_g$ from the question in the given context.
    
    \textit{If this question were asked in this context, I would infer that the speaker is implying this comment.}
    
    \item \textbf{Not prescriptive}: The degree to which the intent is conveyed indirectly, without explicitly stating the intended meaning.
    
    \textit{The context avoids explicitly stating what the speaker intends to convey.}
    
    \item \textbf{Sensible}: The internal coherence and commonsense plausibility of the scenario.
    
    \textit{The context for this question is realistic and follows common sense.}
\end{enumerate}



\subsection{Qualitative Comparison of Human- and LLM-Written Contexts}

Table~\ref{tab:writing} presents representative examples of context–interpretation pairs authored by human annotators and by GPT-4o for the same question–intent combinations. These examples illustrate qualitative differences in how humans and models realize implicature in written contexts, highlighting trade-offs between contextual underspecification and over-explicit interpretation.

\begin{table*}[t!]
\centering
\begin{center}
\setlength{\fboxsep}{4pt}
\colorbox{rowA}{\strut\ \ \ } = Human Writer \hspace{1em}
\colorbox{rowB}{\strut\ \ \ } = GPT-4o
\end{center}

\textbf{Question:} \textit{Are we having dinner at home tonight? } 
\\[6pt]
\centering

\renewcommand{\arraystretch}{1.35}
\rowcolors{2}{rowB}{rowA}
\begin{tabularx}{\textwidth}{m{2cm} X X}
\rowcolor{white}
\textbf{Intent} & \textbf{Context} & \textbf{Implied Comment} \\
\midrule

\multirow{2}{*}{\parbox[c]{2cm}{ \cellcolor{white}\textit{Insult}}}

& A husband asks his wife the day following a failed attempt to cook a nice meal at home &
The cooking is of a low standard. \\

\cellcolor{white}& An adult child asks their parent mockingly after visiting and noticing the state of the messy dining area. &
This place is a mess, and it's clear you can't manage the house. \\
\midrule

\multirow{2}{*}{\centering \cellcolor{white}\textit{Disagree}} 

& A friend suggests staying in for dinner again after eating at home several nights in a row. The other friend, sounding slightly exasperated, asks this question with a raised eyebrow, implying they’d rather go out for a change. &
I don’t want to eat at home again, let’s go out instead. \\

\cellcolor{white}& A roommate questions their fellow roommate after hearing confusing dinner plans, given they had all agreed to order pizza. &
We already planned to not eat at home. \\
\midrule

\end{tabularx}
\textbf{\break \break}
\textbf{Question:} \textit{Can you throw this in the trash? } 
\\[6pt]
\centering

\renewcommand{\arraystretch}{1.35}
\rowcolors{2}{rowA}{rowB}
\begin{tabularx}{\textwidth}{m{2cm} X X}
\midrule

\multirow{2}{*}{\parbox[c]{2cm}{ \cellcolor{white}\textit{encourage}}}
& A parent talking to a young child while doing a task such as cooking in the kitchen. & That the parent is giving a job for the young child to do so that they feel as though they are involved in the task being carried out by the parent and that they are therefore being helpful. \\

\cellcolor{white}& A parent asks their young child in a gentle voice, hoping to build responsibility and independence. &
I believe in you to do this small chore and help out at home. \\

\midrule

{ \cellcolor{white}\textit{}} 

& After sitting through a terrible movie, a film critic sarcastically holds upto the dvd and asks if it should be thrown away clearly mocking how bad the film was. &This is worthless and deserves to be thrown away\\

{ \cellcolor{white}\textit{make\break rhetorical\break commentary}} & A friend asks another after finding a long-expired container in the other's refrigerator, speaking with mock astonishment.	&	This item is clearly trash.\\

\bottomrule
\end{tabularx}
\caption{Examples of context-interpretation pairs authored by human annotators versus GPT-4o}
\label{tab:writing}
\end{table*}

\end{document}